\documentclass[journal]{IEEEtran}
\usepackage{cite}
\ifCLASSINFOpdf
  \usepackage[pdftex]{graphicx}
  \graphicspath{{../Figures/}}
\else
\fi
\usepackage{amsmath}
\usepackage{algorithmic}
\usepackage{array}
  \usepackage[caption=false,font=footnotesize]{subfig}
\usepackage{fixltx2e}
\usepackage{url}
\usepackage{multirow}
\usepackage{multicol}
\usepackage{times}
\usepackage[multidot]{grffile} % allow dots in graphics filenames
\usepackage{verbatim}  % multiline comments via 
\usepackage{mathtools}
\usepackage[usenames,dvipsnames]{xcolor}
\usepackage{color}
\usepackage{ctable}
\usepackage{longtable}
\usepackage[acronym]{glossaries} % required for Acroynyms
\glsdisablehyper % disable hyperlinks for glossary entries
\usepackage{xifthen}

\newcommand{\bi}{\begin{itemize}}
\newcommand{\ei}{\end{itemize}}

\newcommand{\bfig}{\begin{figure}}
\newcommand{\efig}{\end{figure}}

\newcommand{\be}{\begin{equation}}
\newcommand{\ee}{\end{equation}}

\newcommand{\ba}{\begin{eqnarray}}
\newcommand{\ea}{\end{eqnarray}}

\newcommand{\etal}{{et al.}}

\newcommand{\REF}[2][ZZZZ]{\ifthenelse{\equal{#1}{ZZZZ}}% true, no optional arg
  {\index{general}{#2}\ifthenelse{\boolean{draft}}{{\color{red}\it#2}}{#2}}% false, optional arg
  {\index{general}{#1}\ifthenelse{\boolean{draft}}{{\color{red}\it#2}}{#2}}}

\newcommand{\DEF}[2][ZZZZ]{\ifthenelse{\equal{#1}{ZZZZ}}% true, no optional arg
  {\index{general}{#2}\ifthenelse{\boolean{draft}}{{\color{red}\it#2}}{#2}}% false, optional arg
  {\index{general}{#1}\ifthenelse{\boolean{draft}}{{\color{red}\it#2}}{#2}}}

\newcommand{\DEFX}[2][ZZZZ]{\ifthenelse{\equal{#1}{ZZZZ}}% true, no optional arg
  {\index{general}{#2|textbf}\ifthenelse{\boolean{draft}}{{\color{red}\it#2}}{#2}}% false, optional arg
  {\index{general}{#1|textbf}\ifthenelse{\boolean{draft}}{{\color{red}\it#2}}{#2}}}

\newcommand{\model}[1]{\index{code}{#1@\textit{#1}}\ifthenelse{\boolean{draft}}{{\color{green}\Verb+#1+}}{\Verb+#1+}}
\newcommand{\block}[1]{\ifthenelse{\boolean{draft}}{{\color{green}\Verb+#1+}}{\textsf{#1}}}

\newcommand{\func}[2][ZZZZ]{\ifthenelse{\equal{#1}{ZZZZ}}{\index{code}{#2}}{\index{code}{#1}}\ifthenelse{\boolean{draft}}{{\color{green}\Verb+#2+}}{\Verb+#2+}}

\newcommand{\methodb}[2]{\index{code}{#1@\textbf{#1}!.#2}\ifthenelse{\boolean{draft}}{{\color{magenta}\Verb+#1.#2+}}{\Verb+#1.#2+}}
\newcommand{\method}[2]{\index{code}{#1@\textbf{#1}!.#2}\ifthenelse{\boolean{draft}}{{\color{magenta}\Verb+#2+}}{\Verb+#2+}}
\newcommand{\class}[1]{\index{code}{#1@\textbf{#1}}\ifthenelse{\boolean{draft}}{{\color{cyan}\Verb+#1+}}{\Verb+#1+}}
\newcommand{\property}[1]{\index{property}{#1}\ifthenelse{\boolean{draft}}{{\color{cyan}\Verb+#1+}}{\Verb+#1+}}

\newcommand{\presup}[1]{\,{}^{\scriptscriptstyle #1}\!}

\newcommand{\pose}[1][ZZZZ]{\ifthenelse{\equal{#1}{ZZZZ}}{}{\presup{#1}}{\mathbf{\xi}}}

\usepackage{mathrsfs}

\newcommand{\poser}[1][ZZZZ]{\ifthenelse{\equal{#1}{ZZZZ}}{}{\presup{#1}}{\mathscr{R}}}
\newcommand{\poserx}[1][ZZZZ]{\ifthenelse{\equal{#1}{ZZZZ}}{}{\presup{#1}}{\mathscr{R}_x}}
\newcommand{\posery}[1][ZZZZ]{\ifthenelse{\equal{#1}{ZZZZ}}{}{\presup{#1}}{\mathscr{R}_y}}
\newcommand{\poserz}[1][ZZZZ]{\ifthenelse{\equal{#1}{ZZZZ}}{}{\presup{#1}}{\mathscr{R}_z}}
\newcommand{\poserw}[1][ZZZZ]{\ifthenelse{\equal{#1}{ZZZZ}}{}{\presup{#1}}{\mathscr{R}_\omega}}
\newcommand{\poset}[1][ZZZZ]{\ifthenelse{\equal{#1}{ZZZZ}}{}{\presup{#1}}{\mathscr{T}}}
\newcommand{\posetx}[1][ZZZZ]{\ifthenelse{\equal{#1}{ZZZZ}}{}{\presup{#1}}{\mathscr{T}_x}}
\newcommand{\posety}[1][ZZZZ]{\ifthenelse{\equal{#1}{ZZZZ}}{}{\presup{#1}}{\mathscr{T}_y}}
\newcommand{\posetz}[1][ZZZZ]{\ifthenelse{\equal{#1}{ZZZZ}}{}{\presup{#1}}{\mathscr{T}_z}}
\newcommand{\posett}[1][ZZZZ]{\ifthenelse{\equal{#1}{ZZZZ}}{}{\presup{#1}}{\mathscr{T}\!}}

\newcommand{\poseri}[1][ZZZZ]{\ifthenelse{\equal{#1}{ZZZZ}}{}{\presup{#1}}{\mathscr{R}_i}}
\newcommand{\poseti}[1][ZZZZ]{\ifthenelse{\equal{#1}{ZZZZ}}{}{\presup{#1}}{\mathscr{T}_i}}

\newcommand{\twist}[1][ZZZZ]{\ifthenelse{\equal{#1}{ZZZZ}}{}{\presup{#1}}{S}}

\newcommand{\estpose}[1][ZZZZ]{\ifthenelse{\equal{#1}{ZZZZ}}{}{\presup{#1}}{\mathbf{\hat{\xi}}}}
\newcommand{\hpose}[1][ZZZZ]{\ifthenelse{\equal{#1}{ZZZZ}}{}{\presup{#1}}{\hat{\mathbf{\xi}}}}
\newcommand{\posedot}[1][ZZZZ]{\ifthenelse{\equal{#1}{ZZZZ}}{}{\presup{#1}}{\mathbf{\nu}}}
\newcommand{\q}[1][ZZZZ]{\ifthenelse{\equal{#1}{ZZZZ}}{}{\presup{#1}}{\mathring{q}}}

\DeclareMathAlphabet{\mathitbf}{OML}{cmm}{b}{it}
\renewcommand{\vec}[2][ZZZZ]{\ifthenelse{\equal{#1}{ZZZZ}}{}{\presup{#1}}{\mathitbf{#2}}}

\newcommand{\hvec}[2][ZZZZ]{\ifthenelse{\equal{#1}{ZZZZ}}{}{\presup{#1}}{\hat{\vec{#2}}}}
\newcommand{\ovec}[2][ZZZZ]{\ifthenelse{\equal{#1}{ZZZZ}}{}{\presup{#1}}{\mathring{\vec{#2}}}}
\newcommand{\tvec}[2][ZZZZ]{\ifthenelse{\equal{#1}{ZZZZ}}{}{\presup{#1}}{\tilde{\vec{#2}}}}
\newcommand{\evec}[2][ZZZZ]{\ifthenelse{\equal{#1}{ZZZZ}}{}{\presup{#1}}{\hat{\vec{#2}}}}
\newcommand{\dvec}[2][ZZZZ]{\ifthenelse{\equal{#1}{ZZZZ}}{}{\presup{#1}}{\dot{\vec{#2}}}}
\newcommand{\ddvec}[2][ZZZZ]{\ifthenelse{\equal{#1}{ZZZZ}}{}{\presup{#1}}{\ddot{\vec{#2}}}}
\newcommand{\vech}[2][ZZZZ]{\ifthenelse{\equal{#1}{ZZZZ}}{}{\presup{#1}}{\mathitbf{\tilde{#2}}}}
\newcommand{\vecb}[2][ZZZZ]{\ifthenelse{\equal{#1}{ZZZZ}}{}{\presup{#1}}{\bar{\underline #2}}}
\newcommand{\mat}[2][ZZZZ]{\ifthenelse{\equal{#1}{ZZZZ}}{}{\presup{#1}\,}{{\boldsymbol #2}}}
\newcommand{\hmat}[2][ZZZZ]{\ifthenelse{\equal{#1}{ZZZZ}}{}{\presup{#1}\,}{{\hat{\boldsymbol #2}}}}
\newcommand{\dmat}[2][ZZZZ]{\ifthenelse{\equal{#1}{ZZZZ}}{}{\presup{#1}\,}{\dot{\boldsymbol #2}}}
\newcommand{\emat}[2][ZZZZ]{\ifthenelse{\equal{#1}{ZZZZ}}{}{\presup{#1}\,}{\hat{\boldsymbol #2}}}
\newcommand{\matfn}[3][ZZZZ]{\ifthenelse{\equal{#1}{ZZZZ}}{}{\presup{#1}}{{\mat{#2}}\left(#3\right)}}
\newcommand{\Rt}[2][ZZZZ]{\ifthenelse{\equal{#1}{ZZZZ}}{}{\presup{#1}}{{\bf R}\left(#2\right)}}

\newcommand{\point}[2][ZZZZ]{\ifthenelse{\equal{#1}{ZZZZ}}{}{\presup{#1}}{\mathbf{\mathrm{#2}}}}
\renewcommand{\frame}[3][ZZZZ]{\ifthenelse{\equal{#1}{ZZZZ}}{}{\presup{#1}}{\mat{#2}}_{#3}}
\newcommand{\frameh}[3][ZZZZ]{\ifthenelse{\equal{#1}{ZZZZ}}{}{\presup{#1}}{\hat{#2}}_{#3}}
\newcommand{\frameb}[3][ZZZZ]{\ifthenelse{\equal{#1}{ZZZZ}}{}{\presup{#1}}{\bar{#2}}_{#3}}

\newcommand{\pnt}[2][ZZZZ]{\ifthenelse{\equal{#1}{ZZZZ}}{}{\presup{#1}}{\mathbf{#2}}}
\newfont{\School}{pncr}
\newfont{\eightTR}{pncr at 8pt}
\usepackage{bm} % somewhat more comprehensive bolding with \bm

\DeclareMathSymbol{\Gamma}{\mathalpha}{letters}{"00}
\DeclareMathSymbol{\Delta}{\mathalpha}{letters}{"01}
\DeclareMathSymbol{\Theta}{\mathalpha}{letters}{"02}
\DeclareMathSymbol{\Lambda}{\mathalpha}{letters}{"03}
\DeclareMathSymbol{\Xi}{\mathalpha}{letters}{"04}
\DeclareMathSymbol{\Pi}{\mathalpha}{letters}{"05}
\DeclareMathSymbol{\Sigma}{\mathalpha}{letters}{"06}
\DeclareMathSymbol{\Upsilon}{\mathalpha}{letters}{"07}
\DeclareMathSymbol{\Phi}{\mathalpha}{letters}{"08}
\DeclareMathSymbol{\Psi}{\mathalpha}{letters}{"09}
\DeclareMathSymbol{\Omega}{\mathalpha}{letters}{"0A}

\newcommand{\vect}[1]{\bm{#1}}

\newcommand\negative[1]{{\text{-}#1}}
\newcommand\sub[1]{_{\scriptscriptstyle\mathit{#1}}}

\DeclareSymbolFont{EUr}{U}{eur}{m}{n}
\DeclareSymbolFont{EUb}{U}{eur}{b}{n}
\DeclareMathSymbol{\upGamma}{\mathord}{EUr}{"00}
\DeclareMathSymbol{\upDelta}{\mathord}{EUr}{"01}
\DeclareMathSymbol{\upTheta}{\mathord}{EUr}{"02}
\DeclareMathSymbol{\upLambda}{\mathord}{EUr}{"03}
\DeclareMathSymbol{\upXi}{\mathord}{EUr}{"04}
\DeclareMathSymbol{\upPi}{\mathord}{EUr}{"05}
\DeclareMathSymbol{\upSigma}{\mathord}{EUr}{"06}
\DeclareMathSymbol{\upUpsilon}{\mathord}{EUr}{"07}
\DeclareMathSymbol{\upPhi}{\mathord}{EUr}{"08}
\DeclareMathSymbol{\upPsi}{\mathord}{EUr}{"09}
\DeclareMathSymbol{\upOmega}{\mathord}{EUr}{"0A}
\DeclareMathSymbol{\upalpha}{\mathord}{EUr}{"0B}
\DeclareMathSymbol{\upbeta}{\mathord}{EUr}{"0C}
\DeclareMathSymbol{\upgamma}{\mathord}{EUr}{"0D}
\DeclareMathSymbol{\updelta}{\mathord}{EUr}{"0E}
\DeclareMathSymbol{\upepsilon}{\mathord}{EUr}{"0F}
\DeclareMathSymbol{\upzeta}{\mathord}{EUr}{"10}
\DeclareMathSymbol{\upeta}{\mathord}{EUr}{"11}
\DeclareMathSymbol{\uptheta}{\mathord}{EUr}{"12}
\DeclareMathSymbol{\upiota}{\mathord}{EUr}{"13}
\DeclareMathSymbol{\upkappa}{\mathord}{EUr}{"14}
\DeclareMathSymbol{\uplambda}{\mathord}{EUr}{"15}
\DeclareMathSymbol{\upmu}{\mathord}{EUr}{"16}
\DeclareMathSymbol{\upnu}{\mathord}{EUr}{"17}
\DeclareMathSymbol{\upxi}{\mathord}{EUr}{"18}
\DeclareMathSymbol{\uppi}{\mathord}{EUr}{"19}
\DeclareMathSymbol{\uprho}{\mathord}{EUr}{"1A}
\DeclareMathSymbol{\upsigma}{\mathord}{EUr}{"1B}
\DeclareMathSymbol{\uptau}{\mathord}{EUr}{"1C}
\DeclareMathSymbol{\upupsilon}{\mathord}{EUr}{"1D}
\DeclareMathSymbol{\upphi}{\mathord}{EUr}{"1E}
\DeclareMathSymbol{\upchi}{\mathord}{EUr}{"1F}
\DeclareMathSymbol{\uppsi}{\mathord}{EUr}{"20}
\DeclareMathSymbol{\upomega}{\mathord}{EUr}{"21}
\DeclareMathSymbol{\upvarepsilon}{\mathord}{EUr}{"22}
\DeclareMathSymbol{\upvartheta}{\mathord}{EUr}{"23}
\DeclareMathSymbol{\upvaromega}{\mathord}{EUr}{"24}
\DeclareMathSymbol{\upvarphi}{\mathord}{EUr}{"27}
\DeclareMathSymbol{\UpGamma}{\mathord}{EUb}{"00}
\DeclareMathSymbol{\UpDelta}{\mathord}{EUb}{"01}
\DeclareMathSymbol{\UpTheta}{\mathord}{EUb}{"02}
\DeclareMathSymbol{\UpLambda}{\mathord}{EUb}{"03}
\DeclareMathSymbol{\UpXi}{\mathord}{EUb}{"04}
\DeclareMathSymbol{\UpPi}{\mathord}{EUb}{"05}
\DeclareMathSymbol{\UpSigma}{\mathord}{EUb}{"06}
\DeclareMathSymbol{\UpUpsilon}{\mathord}{EUb}{"07}
\DeclareMathSymbol{\UpPhi}{\mathord}{EUb}{"08}
\DeclareMathSymbol{\UpPsi}{\mathord}{EUb}{"09}
\DeclareMathSymbol{\UpOmega}{\mathord}{EUb}{"0A}
\DeclareMathSymbol{\Upalpha}{\mathord}{EUb}{"0B}
\DeclareMathSymbol{\Upbeta}{\mathord}{EUb}{"0C}
\DeclareMathSymbol{\Upgamma}{\mathord}{EUb}{"0D}
\DeclareMathSymbol{\Updelta}{\mathord}{EUb}{"0E}
\DeclareMathSymbol{\Upepsilon}{\mathord}{EUb}{"0F}
\DeclareMathSymbol{\Upzeta}{\mathord}{EUb}{"10}
\DeclareMathSymbol{\Upeta}{\mathord}{EUb}{"11}
\DeclareMathSymbol{\Uptheta}{\mathord}{EUb}{"12}
\DeclareMathSymbol{\Upiota}{\mathord}{EUb}{"13}
\DeclareMathSymbol{\Upkappa}{\mathord}{EUb}{"14}
\DeclareMathSymbol{\Uplambda}{\mathord}{EUb}{"15}
\DeclareMathSymbol{\Upmu}{\mathord}{EUb}{"16}
\DeclareMathSymbol{\Upnu}{\mathord}{EUb}{"17}
\DeclareMathSymbol{\Upxi}{\mathord}{EUb}{"18}
\DeclareMathSymbol{\Uppi}{\mathord}{EUb}{"19}
\DeclareMathSymbol{\Uprho}{\mathord}{EUb}{"1A}
\DeclareMathSymbol{\Upsigma}{\mathord}{EUb}{"1B}
\DeclareMathSymbol{\Uptau}{\mathord}{EUb}{"1C}
\DeclareMathSymbol{\Upupsilon}{\mathord}{EUb}{"1D}
\DeclareMathSymbol{\Upphi}{\mathord}{EUb}{"1E}
\DeclareMathSymbol{\Upchi}{\mathord}{EUb}{"1F}
\DeclareMathSymbol{\Uppsi}{\mathord}{EUb}{"20}
\DeclareMathSymbol{\Upomega}{\mathord}{EUb}{"21}
\DeclareMathSymbol{\Upvarepsilon}{\mathord}{EUb}{"22}
\DeclareMathSymbol{\Upvartheta}{\mathord}{EUb}{"23}
\DeclareMathSymbol{\Upvaromega}{\mathord}{EUb}{"24}
\DeclareMathSymbol{\Upvarphi}{\mathord}{EUb}{"27}

\newcount\ppnum
\newcommand\ppnumber[1]{%
        \ppnum=#1\relax
        \ifnum\ppnum<0
                $-$%
                \ppnum=-\ppnum
        \fi
        \let\pptemp\empty
        \loop\ifnum\ppnum>999
                \count255=\ppnum
                \divide\ppnum by1000
                \count255=\numexpr \count255 - 1000*\ppnum \relax
                \edef\pptemp{,\!\ifnum\count255<100 0\ifnum\count255<10 0\fi\fi
                             \the\count255 \pptemp}%
        \repeat
        \the\ppnum
        \pptemp
}

\newacronym{ACFR}{ACFR}{Australian centre for field robotics}
\newacronym{ACRV}{ACRV}{Australian centre for robotic vision}

\newacronym{AUV}{AUV}{autonomous underwater vehicle}
\newacronym{UAV}{UAV}{unmanned aerial vehicle}
\newacronym{USV}{USV}{unmanned surface vehicle}
\newacronym{UGV}{UGV}{unmanned ground vehicle}
\newacronym{GPS}{GPS}{global positioning system}
\newacronym{SLAM}{SLAM}{simultaneous localisation and mapping}
\newacronym{SfM}{SfM}{structure-from-motion}
\newacronym{RO}{refractive object}{refractive object}
\newacronym{RLFF}{RLFF}{refracted light field feature}
\newacronym{EPI}{EPI}{epipolar planar image}

\newacronym{MDSP}{MDSP}{multi-dimensional signal processing}
\newacronym{ROS}{ROS}{region of support}
\newacronym{DOF}{DOF}{degree-of-freedom}
\newacronym{RMS}{RMS}{root mean square}
\newacronym{SNR}{SNR}{signal-to-noise ratio}
\newacronym{CNR}{CNR}{contrast-to-noise ratio}
\newacronym{PCA}{PCA}{principal component analysis}

\newacronym{FIR}{FIR}{finite impulse response}
\newacronym{IIR}{IIR}{infinite impulse response}
\newacronym{DFT}{DFT}{discrete Fourier transform}
\newacronym{FFT}{FFT}{fast Fourier transform}
\newacronym{PSNR}{PSNR}{peak signal-to-noise ratio}
\newacronym{FPGA}{FPGA}{field programmable gate array}
\newacronym{GPU}{GPU}{graphics processing unit}
\newacronym{ASIC}{ASIC}{application-specific integrated circuit}
\newacronym{BW}{BW}{bandwidth}

\newacronym{PSF}{PSF}{point spread function}
\newacronym{FOV}{FOV}{field of view}
\newacronym{BRDF}{BRDF}{bidirectional reflectance distribution function}
\newacronym{FWHM}{FWHM}{full width at half maximum}

\newacronym{RANSAC}{RANSAC}{random sampling and consensus}
\newacronym{IBVS}{IBVS}{image-based visual servoing}
\newacronym{PBVS}{PBVS}{position-based visual servoing}
\newacronym{VS}{VS}{visual servoing}
\newacronym{LF}{LF}{light field}
\newacronym{LF-IBVS}{LF-IBVS}{light field image-based visual servoing}
\newacronym{M-IBVS}{M-IBVS}{monocular image-based visual servoing}
\makeglossaries
\glsunset{RO}  % prevents RO from displaying in full/long form in first appearance - behaves more like a variable now

\begin{document}
\title{Refractive Light-Field Features for Curved Transparent Objects in Structure from Motion}

\author{Dorian Tsai$^{1}$, Peter Corke$^{1}$, Thierry Peynot$^{1}$, Donald G. Dansereau$^{2}$% <-this % stops a space
% \thanks{Submitted 31 January 2018}
\thanks{This research was partly supported by the Australian Research Council (ARC) Centre of Excellence for Robotic Vision (CE140100016).
%We also thank the other members of the ACRV for their insight and guidance.
}
\thanks{$^{1}$D. Tsai, T. Peynot and P. Corke are with the Australian Centre for Robotic Vision, Queensland University of Technology (QUT), Brisbane, Australia {\tt\small \{dy.tsai\}@qut.edu.au}}
\thanks{$^{2}$D. Dansereau is with the Sydney Institute for Robotics and Intelligent Systems, University of Sydney {\tt\small donald.dansereau@sydney.edu.au}}
% \thanks{Manuscript received July 30, 2020; revised August 26, 2020.}
}

\markboth{Journal of Robotics \& Automation Letters}%
{Tsai \MakeLowercase{\textit{et al.}}: Refracted Light-Field Features}

\maketitle

\begin{abstract}
Curved refractive objects are common in the human environment, and have a complex visual appearance that can cause robotic vision algorithms to fail. Light-field cameras allow us to address this challenge by capturing the view-dependent appearance of such objects in a single exposure. We propose a novel image feature for light fields that detects and describes the patterns of light refracted through curved transparent objects. We derive characteristic points based on these features allowing them to be used in place of conventional 2D features. Using our features, we demonstrate improved structure-from-motion performance in challenging scenes containing refractive objects, including quantitative evaluations that show improved camera pose estimates and 3D reconstructions. Additionally, our methods converge 15-35\% more frequently than the state-of-the-art.
% Importantly, we show that a large proportion of scenes that were previously prohibitively challenging are now tractable using our approach. 
Our method is a critical step towards allowing robots to operate around refractive objects, with applications in manufacturing, quality assurance, pick-and-place, and domestic robots working with acrylic, glass and other transparent materials.
\end{abstract}

%\begin{IEEEkeywords}
%Computational imaging, light-fields, robot perception, structure from motion
%\end{IEEEkeywords}

% \fi
\IEEEpeerreviewmaketitle

%!TEX root = main.tex

\section{Introduction}
Transparent, or \glspl{RO} are often found in urban settings and industrial applications. However, many robotic vision algorithms find these objects particularly difficult to perceive. 
%The assumption of Lambertianness  % Peter and Thierry have never encountered the word. Given that they are likely the target audience, I am removing the term "Lambertianness"
Assuming a Lambertian surface---that the appearance of a point on an object does not change with viewpoint---is common, but \glspl{RO} violate this assumption. Their appearance from a particular camera pose is a distorted view of the scene behind them. Thus points on the object's surface can change dramatically in appearance with small changes in viewpoint. Consequently, robotic vision algorithms, including most approaches to \gls{SfM} and \gls{SLAM}, perform poorly around \glspl{RO}. These algorithms yield incorrect camera trajectories and 3D shape estimates and sometimes fail to converge~\cite{ihrke2010transparentsurvey, tsai2018refractedfeatures, xu2015transparentobjectsegmentation}.

\begin{figure}[t!]
    \centering
    \subfloat[][]{\includegraphics[width=0.95\columnwidth]{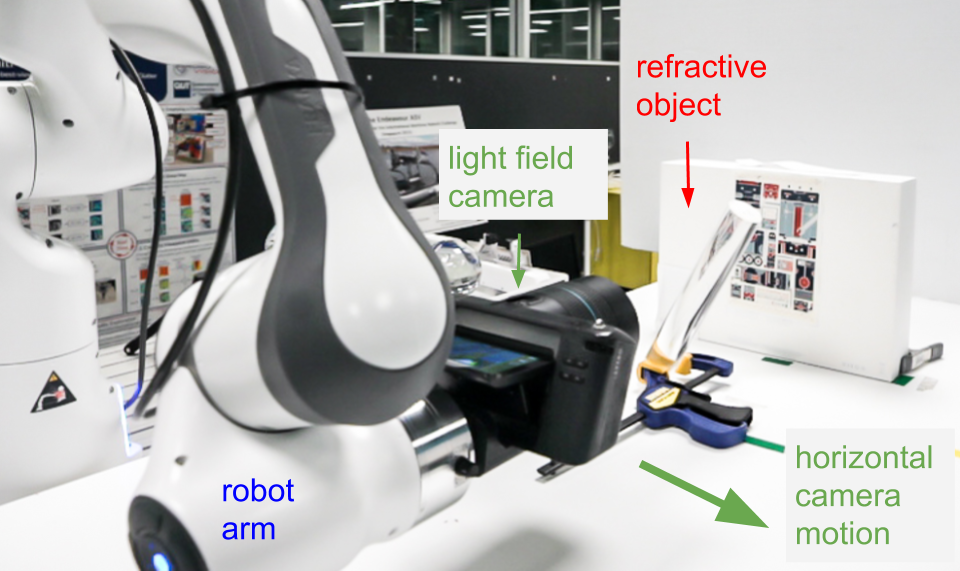} % rlffcover.png
  \label{fig:robotview}}\\
    \subfloat[][]{\includegraphics[,width=0.475\columnwidth]{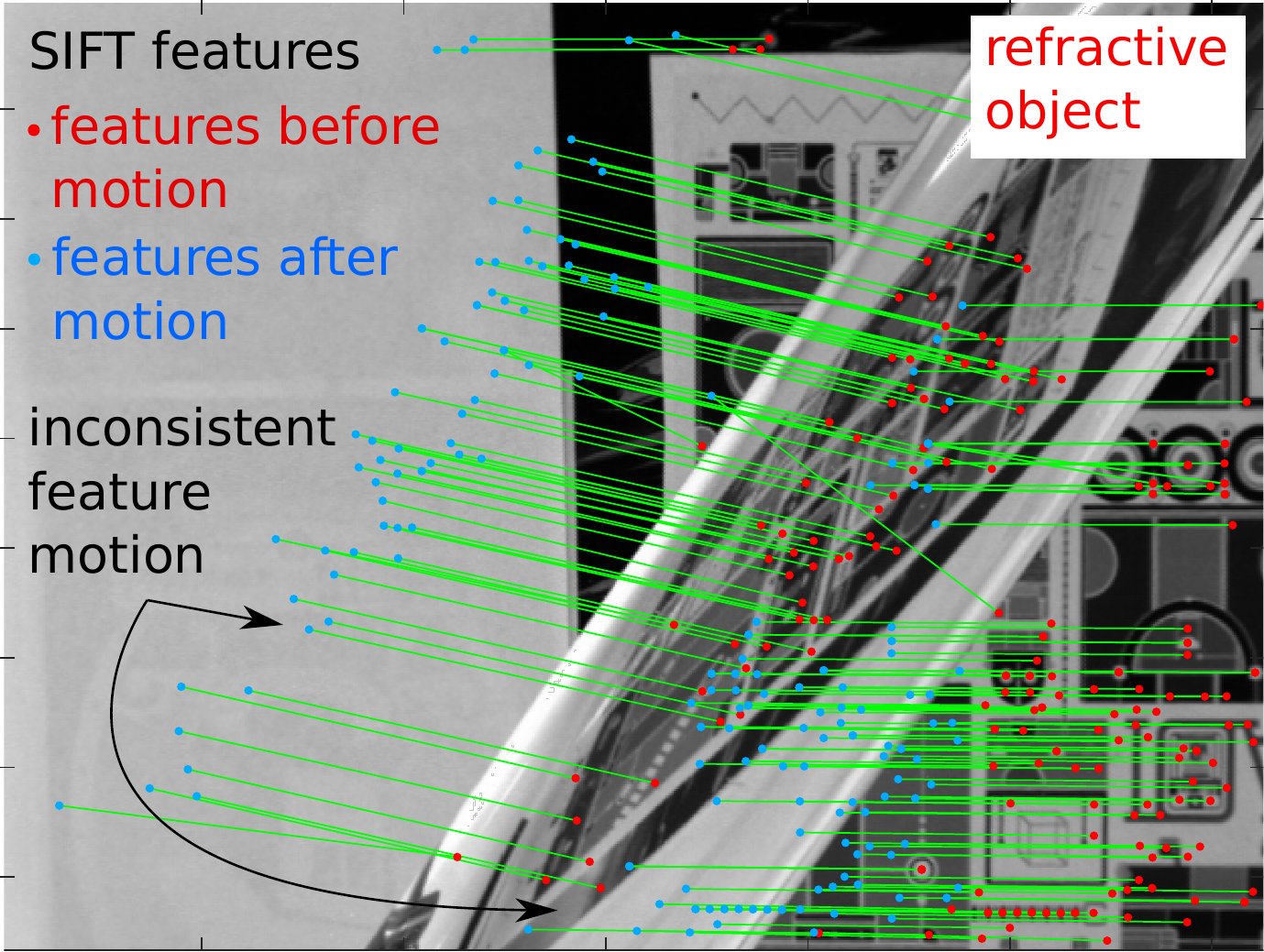} 
    %\subfloat[][]{\includegraphics[,width=0.475\columnwidth]{Figures/IMG_6699__Decoded_2DSIFT_feature_motion.pdf} 
  \label{fig:2d_sift_ftr_motion}}\hfil
  \subfloat[][]{\includegraphics[width=0.475\columnwidth]{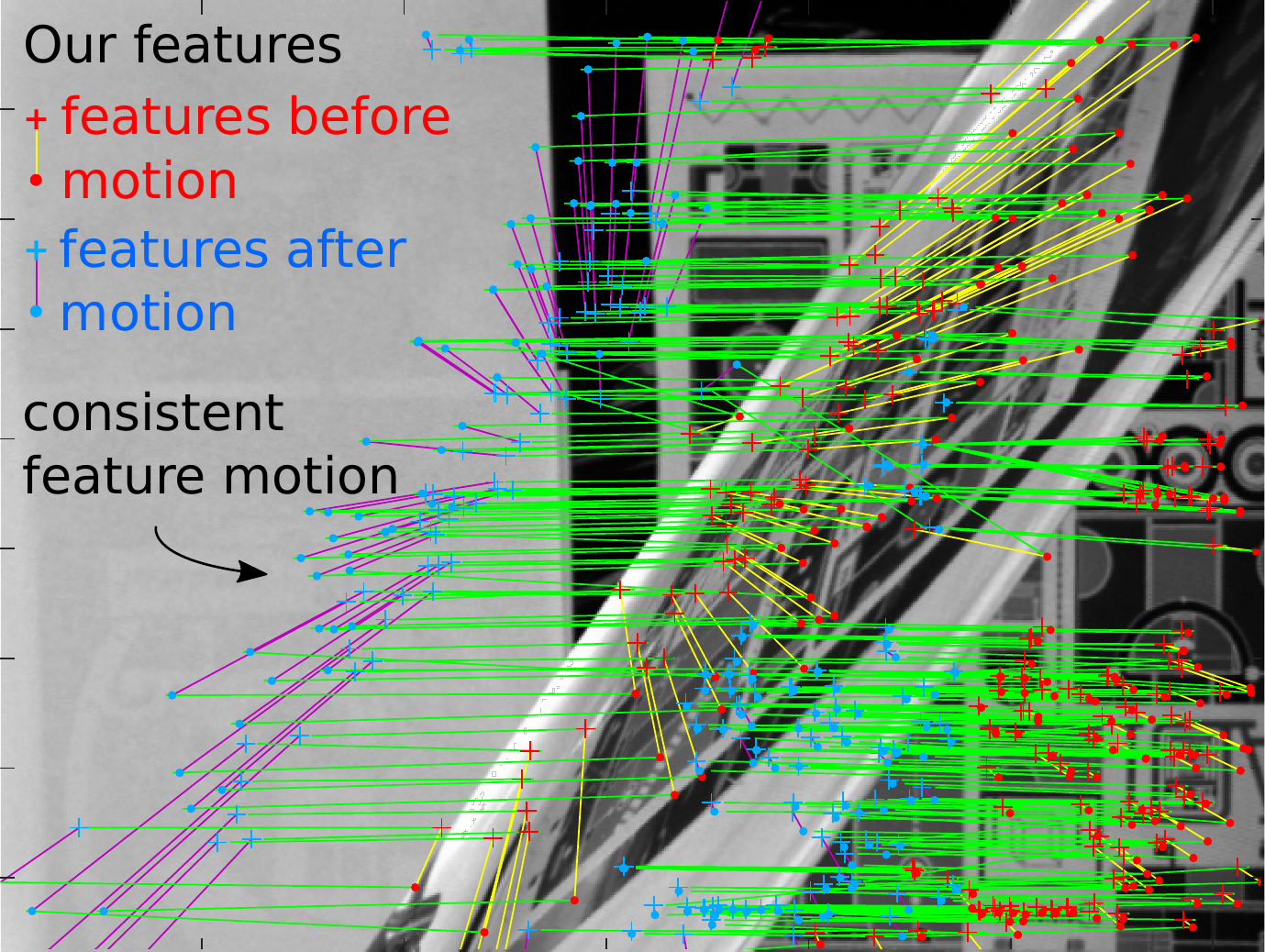}
\label{fig:synth_stereo_rlff_ftr_motion}}
% \subfloat[][]{\includegraphics[width=0.475\columnwidth]{Figures/IMG_6699__Decoded_synthStereoRLFF_feature_motion.pdf}
% \label{fig:synth_stereo_rlff_ftr_motion}}
    \caption{Comparison of conventional and proposed image features: (a) a robot-mounted LF camera moves horizontally while observing an acrylic cylinder; (b) as seen by the camera, 2D SIFT features' apparent motion (green) between two frames (from red to blue) is inconsistent across the scene, due to distortion through the refractive object (causing the vertical shift in the refracted image features); (c) the proposed feature exhibits consistent apparent motion (green) between views (only exhibiting horizontal image feature motion), enabling structure-from-motion to operate correctly. The proposed approach yields two characteristic points for refracted features, simplifying to a single point for Lambertian features.}
    \label{fig:title}
\end{figure}

%For instance, in robotic vision, image matching is used to integrate data from different viewpoints and features often play a central role in this process. \textit{Features} are distinct aspects of the scene that can be reliably and repeatedly identified from different viewpoints and/or across different viewing conditions. \textit{Image features} are those features recorded in the image as a set of pixels by the camera that can be automatically detected and extracted as a vector of numbers, which is referred to as an image feature vector. \textit{Image feature vectors} abstract raw and dense image information into a simpler, smaller and more compact relevant representation of the data. 
% introduce SfM context:
%In \gls{SfM}, both scene structure and camera pose are recovered from 2D images and is widely applicable to many computer and robotic systems. Image features are typically used for image registration and serve as a basis for the entire \gls{SfM} pipeline. For an eye-in-hand robot attempting to interact with an \gls{RO}, such as  Fig.~\ref{fig:title}, image features visible through the object move differently from the rest of the scene, which can result in erratic pose estimates that can negatively affect system performance.   

% define image features
In this work we propose a new feature detector for \gls{LF} cameras that allows existing feature-based algorithms to operate in scenes dominated by \glspl{RO}. \textit{Image features} are distinct points of interest that can be repeatedly and reliably identified from different viewpoints. These form the basis for a range of robotic perception tasks including visual odometry and 3D reconstruction via \gls{SfM} and \gls{SLAM}~\cite{corke2013roboticvision}. When image features are visible through an \gls{RO} they exhibit apparent motion inconsistent with the scene geometry and camera trajectory, an example is seen in Fig.~\ref{fig:2d_sift_ftr_motion}. We refer to these as \textit{refracted image features}. Their view-dependent nature violates assumptions that underpin conventional vision algorithms, which can prevent them from operating as intended~\cite{ihrke2010transparentsurvey, tsai2018refractedfeatures}.
% distorted view-dependent phenomenon as refracted image features, which can prevent conventional vision algorithms from operating~\cite{tsai2018refractedfeatures}. 

\Gls{LF} cameras are an emerging technology that captures dense and uniformly-sampled multiple views of the scene in one exposure. A single \gls{LF} image describes view-dependent effects such as occlusion, specular reflection, and in particular, refraction. We exploit this in the context of \gls{SfM} in a step towards more reliable operation around \glspl{RO}, such as an eye-in-hand robot grasping a transparent wine glass. 
We propose a method of detecting refracted image features based on the patterns of light passing through curved \glspl{RO}. From these we extract characteristic points with more consistent apparent motion, and show that these can directly enable feature-based algorithms like \gls{SfM} to operate around \glspl{RO}.
% TODO: motivate example where robot/eye-in-hand grasping scenario where scene is dominated by a refractive object

Although previous work has shown \gls{LF} capture offers advantages in detecting reliable features~\cite{dansereau2019liff} and ignoring refracted features~\cite{tsai2018refractedfeatures}, none has to our knowledge described the detection and use of refracted features for robotic applications.  We thus propose a novel feature detector for \glspl{RO}: the \gls{RLFF}. Whereas conventional features identify patterns in the geometry or texture of objects, the \gls{RLFF} considers the structure of light refracted by \glspl{RO}, finding characteristic points in the free space between objects.

Our key contributions are:
\begin{itemize}
	\item we describe a new kind of feature, the \gls{RLFF}, that exists in the patterns of light refracted through objects;
	\item we propose efficient methods for detecting and extracting \gls{RLFF} features from \gls{LF} imagery, and for describing them in terms of characteristic points that can be employed in place of conventional features like SIFT; and 
    \item we demonstrate that using \glspl{RLFF} improves \gls{SfM} performance in scenes dominated by \glspl{RO}, yielding more accurate camera trajectory estimates, 3D reconstructions, and more robust convergence, even in complex scenes where state-of-the-art methods otherwise fail.
	%\item We propose a compact representation for the RLFF that allows fast detection and, which is based on the local projections of the background through an \gls{RO}. We assume that the surface of the \gls{RO} can be locally approximated as having two (orthogonal) surface curvatures. We can then model the local part of the \gls{RO} as a (toric) astigmatic lens. 
	% which is a LLFF that has been refracted by an RO. This feature is literally defined by light, which is entirely novel in the field of robotics!
	%\item We provide a novel method to extract the RLFF from the LF using Eigenvalue Decomposition, which is a linear process and can use all sub-views of the LF, where  some prior work did not always use the full LF. % have to cite our work that only used the central cross? Not sure if using the full LF is worth noting.
	% \item extension of finding refracted vs Lambertian (a better overall method of detecting refracted features)
	%\item Our code and data are available at (url).
\end{itemize}

To evaluate the \glspl{RLFF} we captured \gls{LF} imagery using a Lytro Illum camera mounted on a robotic arm.  We captured 218 \glspl{LF} of 20 challenging scenes containing a variety of \glspl{RO} and Lambertian objects. The dataset and code associated with this paper can be accessed at \textit{https://tinyurl.com/rlff2021}.

\textbf{Limitations:} 
This work is inspired chiefly by applications dominated by smooth curved objects like drinking glasses and other manufactured transparent items. Evaluation with flat refractive objects is limited, and we expect adaptation of the method may be required.  As with any feature-based method, the presence of some texture is required for the algorithm to work. In particular, \glspl{RLFF} only occur when texture is visible through a \gls{RO}. The approach here will therefore not work with frosted or very complex surfaces through which scene content is not visible. We also assume a geometric ray-based optics approach, so strong defocus effects through highly distorting objects are not considered.

The rest of this paper is organised as follows. We review related work in Section II. In Section III, we discuss the optics of the lens elements that inform the behaviour
of our \gls{RLFF}. Next, the formulation and extraction of our \gls{RLFF} are described in Section IV. Experimental results using our feature in \gls{SfM} and comparison to traditional 2D SIFT features are presented and discussed in Section V. Lastly, in Section VI, we conclude the paper and explore avenues for future work.

%!TEX root = main.tex

\section{Related Work}
\label{sec:related}
% previous work on dealing with RO, features, LF, and how our work is different from theirs/how we fulfill a different research niche

A variety of approaches for detecting and reconstructing \glspl{RO} using vision have been considered in previous work~\cite{ihrke2010transparentsurvey,miyazaki2005polarization}; however, many require known light sources with bulky configurations that make them impractical for mobile robotics. Other vision-based methods allow for robotic manipulation of \glspl{RO}~\cite{choi2012modelbasedglassobjects, walter2015glossy, lysenkov2013recognition, zhou2018plenoptic}; however, they rely on having a 3D model of the object a priori. Complete and accurate 3D models and refractive indices of \glspl{RO} are often difficult, time-consuming and expensive to obtain, or simply not available~\cite{ihrke2010transparentsurvey}. When such information is not available, localization, manipulation and control of and around \glspl{RO} becomes much harder. 
%Choi et al.~developed a method to localise refractive objects in real-time with a monocular camera~\cite{choi2012modelbasedglassobjects}. The contours from a given image were matched to a database of refractive object contours with known poses, and then efficiently searched/matched to a database. Walter et al.~did so with an LF camera combined with an RGB-D sensor~\cite{walter2015glossy}. Lysenkov et al.~recognised and estimated the pose of rigid transparent objects using an RGB-D (structured light) sensor~\cite{lysenkov2013recognition}.
% Recently, Zhou et al.~used an LF camera to recognise and grasp a refractive object by developing a light-field descriptor based on the distribution of depths observed by the LF camera~\cite{}. However, all of these previous works rely on having 

% LF-based approaches to SfM have recently been of interest.
%~\cite{johannsen2015linear, zhang2017ray, nousias2019large}. 
Recently, \gls{LF} cameras have been considered in the \gls{SfM} framework~\cite{johannsen2015linear,nousias2019large}. However, these methods employ conventional 2D image features that occur on 3D surfaces. In this paper, we propose a novel 4D feature that is defined by patterns of light that are not necessarily fixed to the surface of an object. We compare our feature to traditional 2D SIFT features in monocular and stereo-based \gls{SfM}.

	% \item Monocular vision with occluding edges~\cite{ham2017texturefeatures}
% LF features for Lambertian scenes~\cite{teixeira2017epipolarlightfieldsift, tosic2014lightfieldscalespace, dansereau2019liff, tsai2016lfvisualservo}.
For \gls{LF}-specific features, Tosic~\etal~developed a type of \gls{LF}-edge feature~\cite{tosic2014lightfieldscalespace}; however, our interest is in keypoint features, which tend to be more uniquely identifiable, and are more commonly applied to visual servoing and \gls{SfM} tasks. Tsai~\etal~developed the first \gls{LF} image-based visual servoing algorithm that uses a feature combining central-view image coordinates and depth-dependent \gls{LF} slope~\cite{tsai2016lfvisualservo}. 
% I don't think central view is defined yet, might be ambiguous? Not sure if the following sentence is needed
% the next sentence would, I think only confuse someone not already familiar with this; slope is also not familiar;  
%For an \gls{LF} consisting of $n\times n$ grid of 2D sub-images, the central view is the central $(n/2,~n/2)$ sub-image. 
Teixeira~\etal~used \glspl{EPI} to detect reliable Lambertian image features~\cite{teixeira2017epipolarlightfieldsift}. Similarly, Dansereau~\etal~proposed the Light-Field Feature (LiFF) detector and descriptor~\cite{dansereau2019liff}, which focused on detecting and describing reliable Lambertian image features in a scale-invariant manner. However, all of these \gls{LF} features are designed for Lambertian scenes, and are thus not suitable for describing refracted image features.  
	
\glspl{LF} have been considered for \gls{RO} recognition. 
%\ROs~\cite{maeno2013light, xu2015transparentobjectsegmentation, tsai2018refractedfeatures}. 
Maeno~\etal~proposed an \gls{LF} distortion feature (LFD), which modelled an object's refraction pattern as image distortion~\cite{maeno2013light}. %based on differences in the corresponding image points between the multiple views of the \gls{LF}, 
%captured by a large-baseline (relative to the \gls{RO}) \gls{LF} camera array
However, the authors observed  poor recognition performance due to specular reflections and changes in camera pose. Xu~\etal~used the LFD as a basis for \gls{RO} image segmentation~\cite{xu2015transparentobjectsegmentation}. Corresponding image features from all views in the \gls{LF} were fitted to the single normal of a 4D hyperplane using singular value decomposition (SVD). Tsai~\etal~extended this work to show that a 3D point manifests as a plane in 4D that has two orthogonal normal vectors. This was used to distinguish more types of \gls{RO}, with a higher rate of detection in order to reject refracted scene content~\cite{tsai2018refractedfeatures}.
	
In this paper, we propose a novel \gls{RLFF} based on the appearance of background texture through a \gls{RO}. We extend existing theory to derive methods for detecting, extracting and estimating the 4D structure of an \gls{RLFF} in the \gls{LF}.  We use the full \gls{LF} to detect and extract each feature, making maximal use of available information. We employ the proposed \gls{RLFF} to allow \gls{SfM} to operate in scenes dominated by \glspl{RO}. Notably, while prior work focused on detecting and rejecting the refracted scene content, our approach directly uses both the Lambertian and refracted scene content for more reliable camera pose and 3D shape estimation.

%!TEX root = main.tex
\section{Describing Curved Refractive Objects}
\label{sec:optics}

% for the most part, we use astigmatic lens, but when we show specific instatiations of the lens (eg Fig 2, 3), we can say toric lens (similar to RO vs refractive cylinder).

\begin{figure}
  \centering
  \subfloat[][]{\includegraphics[height=0.35\columnwidth]{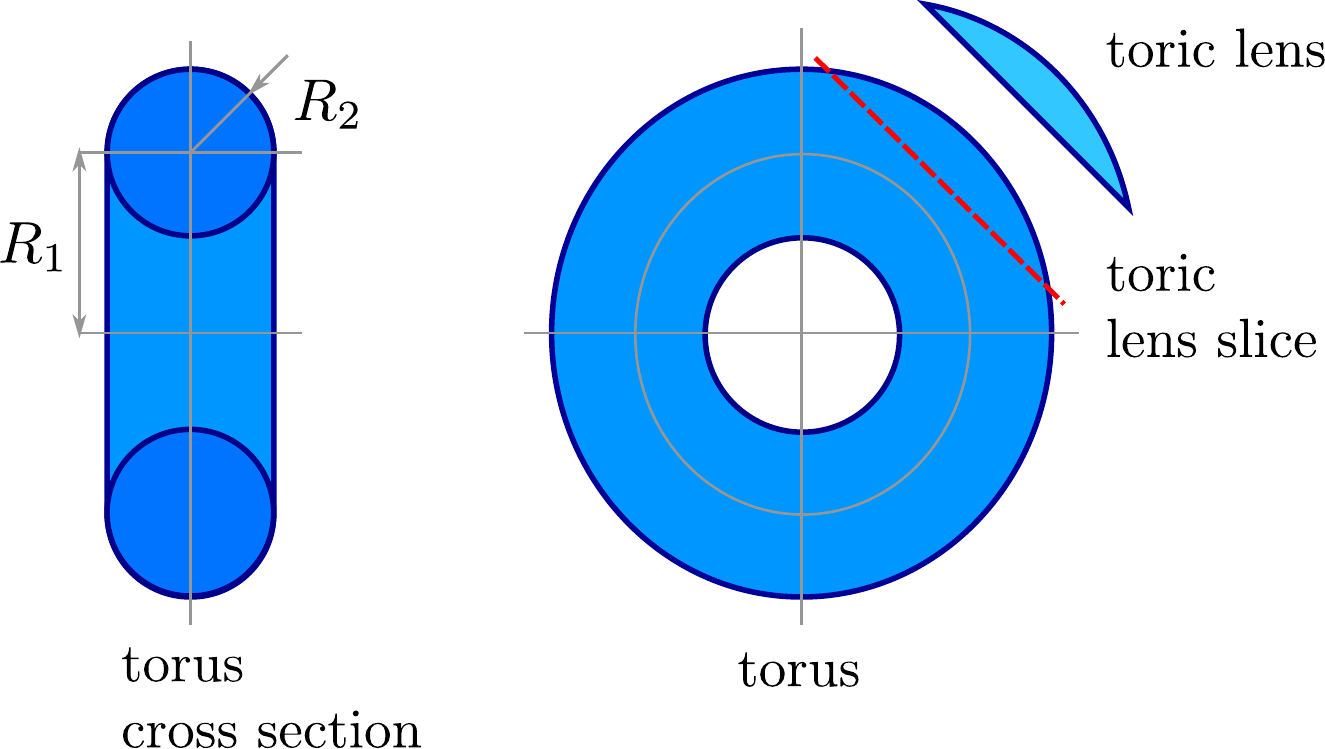}\label{fig:torus}}\hfil 
  \subfloat[][]{\includegraphics[height=0.35\columnwidth]{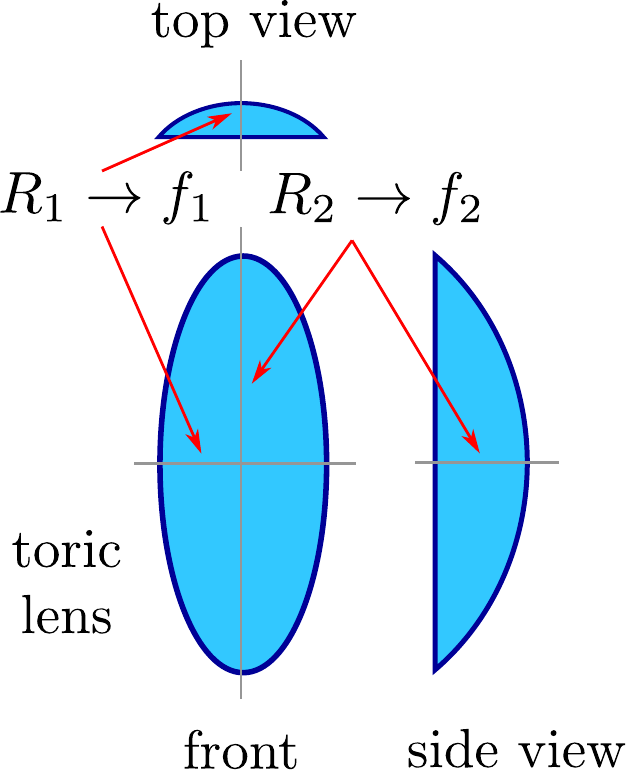}\label{fig:toric}}
  \caption[]{A complex refractive object can be described in terms of the local curvature of small surface patches. (a) The surface of a torus with radii $R_1$ and $R_2$ is sliced (dashed red) to form a toric lens surface; (b)~The lens surface is defined by two local radii of curvature, and will focus light to focal lines at two distinct focal depths. For the more general astigmatic surface, the axes of curvature need not be orthogonal. 
  } 
  \label{fig:torusToToric} 
\end{figure}

We wish to understand the visual appearance of background points imaged through \glspl{RO}, so that we can locally approximate the surface of a \gls{RO} as an astigmatic lens. 
We begin by investigating the behaviour of light as it travels from the background texture and enters the object. Where the light intersects with the object, the object's local surface curvature will determine its path, just as in the case of the surface of a lens. In fact, we can describe the entire \gls{RO} as a collection of surface patches, each distorting light based on its local curvature. 
Fig.~\ref{fig:torusToToric} illustrates this concept for a toric \gls{RO}, a specific case of an astigmatic \gls{RO}. Here we highlight part of the torus' surface that has a local shape well described by two orthogonal axes of curvature, each with a corresponding radius of curvature. 
In the general case of an asymmetric surface, the axes of curvature need not be orthogonal, and the result is an astigmatic surface, part of an astigmatic lens~\cite{hecht2002optics}.  

Other common optical surfaces can be described as special cases of the astigmatic surface. A spherical surface has two identical radii of curvature, and focuses a point source of light to a single point. A cylindrical surface has an infinite radius of curvature in one direction, and focuses a point to a line. 

As light leaves the \gls{RO}, it encounters a second surface and is again distorted. As with the first surface, the behaviour is determined by the local curvature of the object. In the general case in which both entrance and exit surfaces are astigmatic, their combined effect is to behave like an astigmatic lens~\cite{hecht2002optics}. 

\begin{figure}
  \centering 
  \includegraphics[width=0.9\columnwidth]{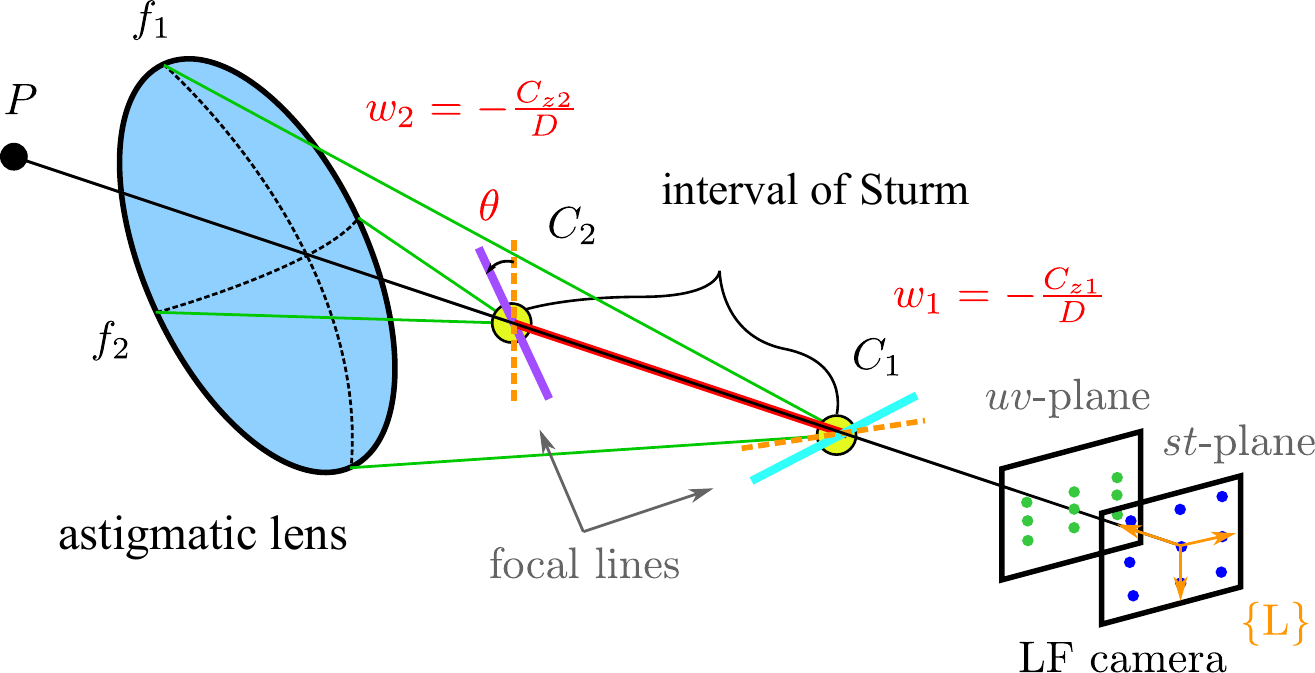}\label{fig:rlff_def}
  \caption{A point $P$ imaged through an astigmatic lens (blue) forms a distorted pencil of rays. Two lines of focus (purple and cyan) form at depths $C_1$ and $C_2$, and the shortest line connecting these is the interval of Sturm (red). The two 3D points $C_1$ and $C_2$ that describe the interval of Sturm that comprise our RLFF are shown (yellow/black circles). Observing the scene with an LF camera, we can estimate the orientations of the focal lines and the endpoints of the interval of Sturm. These phenomena are stable with respect to camera pose, and form the basis for our proposed refractive feature.}
  \label{fig:IntervalOfSturm} 
\end{figure}

Fig.~\ref{fig:IntervalOfSturm} depicts the image of a point as it is seen through a general astigmatic lens. Note there are two focal lines at distinct depths $C_1$ and $C_2$. The shape of the bundle of rays passing through the astigmatic lens is known as an astigmatic pencil. Mathematician Jacques Sturm (1838) investigated the properties of the astigmatic pencil, and it is thus also known as Sturm's conoid~\cite{hecht2002optics}. The shortest line segment connecting the two focal lines is known as the interval of Sturm. 

As a line segment, the interval of Sturm can be described by two 3D points. These points have the desirable properties that they can be observed from different positions in the scene, they do not shift significantly as a function of viewpoint, and they can be estimated from a single \gls{LF} image. These will therefore form the basis for the proposed refractive feature. Though our discussion is concerned with points, generalisation to more complex textural shapes, and in particular the corners or blobs that make up conventional image features, is straightforward.

%The circular cross-section where the pencil has the smallest area is known as the circle of least confusion.  % we never use this concept so no need to introduce it

% The focal line is parallel to the longitudinal axis of the lens. Effectively, the lens compresses the image of the background in the direction perpendicular to the focal line. The background image is unaltered in the direction parallel to the focal line.
%A toric lens has the same optical effect as two perpendicular cylindrical lenses combined. Visually, this is seen as a ``flattening'' of rays with respect to their respective axes at these two distances~\cite{freeman1990optics}. 

% Fig.~\ref{fig:toricLensView} shows a rendering of the visual effect of a toric lens, here shown distorting a blue circle on a checkerboard pattern. The diversity of behaviours is complex and difficult to understand in terms of image geometry. We thus turn to the study of rays as they pass through such a lens. 

% % take-away: 

% specifically, we use the pair of 3D points defined by the endpoints of the inverval of Sturm as our RLFF. 
% the reasoning behind this: while in motion, these points move less with respect to motion than all other features

%!TEX root = main.tex
\section{Refracted Light Field Features}
\label{sec:rlff_ext}
% todo: align notation, remove redundancy, align presentation around toric vs more general astigmatic lenses

\begin{figure} 
  \centering
  \subfloat[][]{\includegraphics[height=0.4\columnwidth]{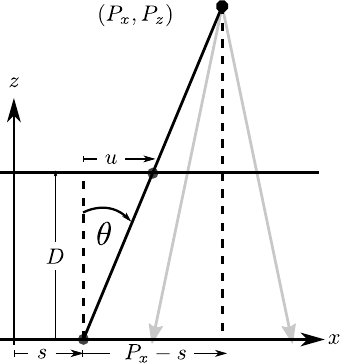}\label{fig:lfDepthROVS}}\hfil 
  \subfloat[][]{\includegraphics[height=0.4\columnwidth]{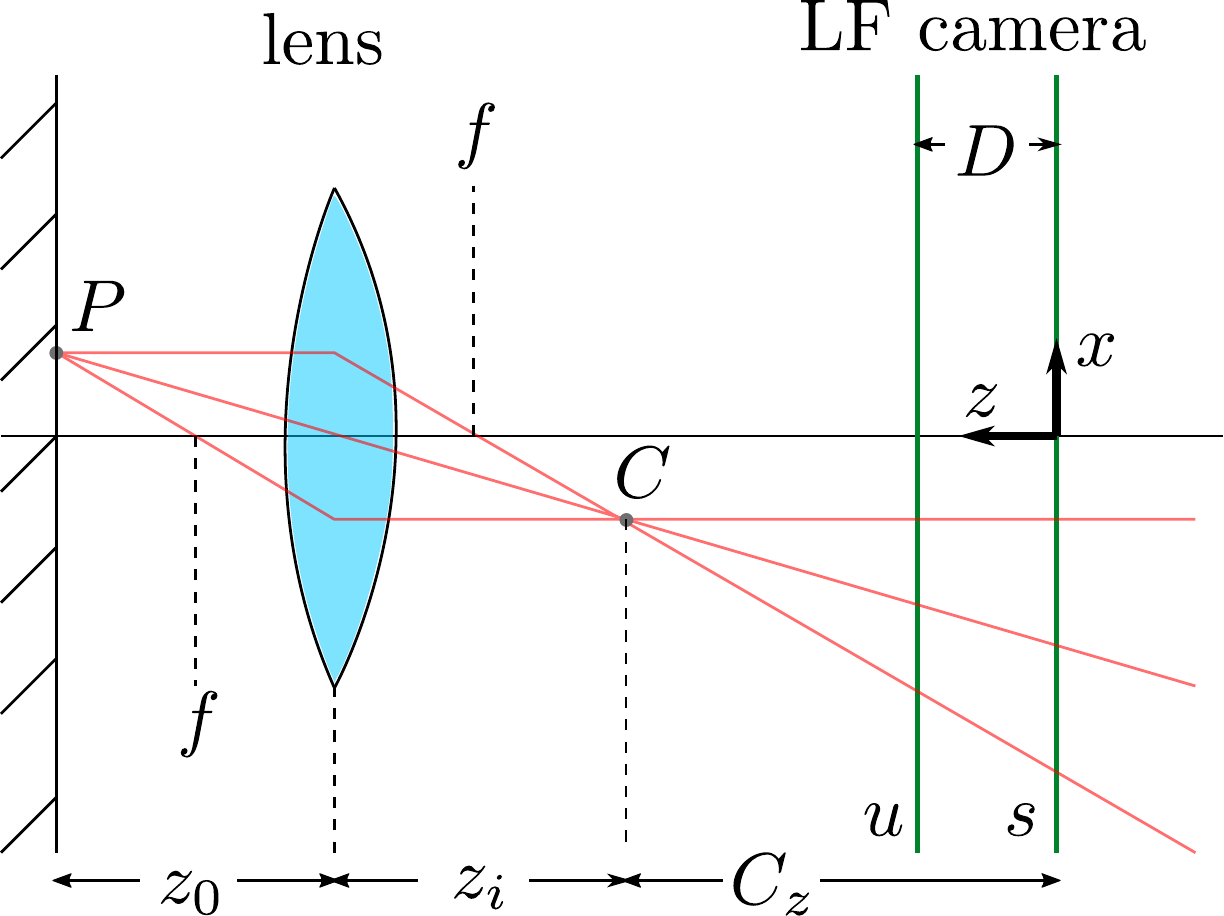}\label{fig:rovsProjection}}
  \caption[]{(a) Geometry of the point-plane correspondence: in a slice of the scene, a Lambertian point $P$ manifests as a line in $s,u$ with slope inversely proportional to $P_z$~\cite{adelson1992single}; the same holds in the $t,v$ dimensions yielding a plane in the 4D LF~\eqref{eq_Hyperfan_PtPlane}. (b) Imaging the point through an axis-aligned astigmatic lens, in a single slice of the \gls{LF}, the point $P$ manifests as a focal line going into the page at $C$, yielding a line in $s,u$-space. In a similar slice (not shown), $P$ appears as a second focal line at a different depth (see Fig.~\ref{fig:IntervalOfSturm}), yielding a line at a different slope in $t,v$-space; the net result is again a plane, but with unequal slopes for the two slices~\eqref{eq_RLFF_GeneralizedPtPlane}.} 
  \label{fig:rovsLensProjections} 
\end{figure} 

Whereas a conventional 2D image feature is defined by a single location in space, e.g.~the position of a textural corner or centroid of a blob, the \gls{RLFF} is more complex, defined by the interval of Sturm, depicted in Fig.~\ref{fig:IntervalOfSturm}. This section describes our method for detecting the \gls{RLFF} and estimating its parameters from a single \gls{LF} exposure. 

We parameterise the \gls{LF} using the relative two-plane parameterisation~\cite{levoy1996light}. A light ray $\phi$ has coordinates $\vec{\phi}=[s,t,u,v]$, where $s,t$ and $u,v$ describe the points of intersection with two reference planes separated by an arbitrary distance $D$, and $s,t$ are chosen to be further from the scene as depicted in Figs.~\ref{fig:IntervalOfSturm},~\ref{fig:rovsLensProjections}. In the relative parameterisation, $u,v$ are expressed relative to $s,t$. In the sampled \gls{LF}, we employ the discrete variables $i,j,k,l$, where $i,j$ select a sub-image, and $k,l$ select a pixel from that image.

\subsection{Feature Detector and Descriptor}

There is an opportunity to exploit the local 4D structures of light refracted through objects to define a unique feature detector and descriptor. However, for this work we are focused chiefly with identifying and extracting the parameters of refractive features. We therefore leverage existing 2D tools to detect and describe \gls{RLFF} features. 
In an approach similar to~\cite{johannsen2015linear} and~\cite{teixeira2017epipolarlightfieldsift}, we apply a SIFT detector to each sub-image of the \gls{LF}, then match these between views. We employ Root SIFT descriptors applied to the central sub-image of the 2D feature for more reliable matching~\cite{arandjelovic2012rootsift}, and match between \gls{LF} sub-images based on the Euclidean distance between descriptors. Only allowing features that match across a minimum number of sub-images allows us to reject spurious detections. In future work, we envision extending a direct \gls{LF} feature detector like LiFF, which detects features by simultaneously considering the entire \gls{LF}~\cite{dansereau2019liff}. 
% We employ the SIFT descriptor applied to the central sub-image of the 2D feature. 
%To detect features we applied 2D SIFT to every \gls{LF} sub-image~\cite{lowe2004distinctive} as in~\cite{teixeira2017epipolarlightfieldsift}. 
 % This yielded a set of potential \gls{RLFF}s. 

The detection process yields a set of discrete-space observations $\vect{n}$, each of the form $\vect{n_i} = [i,j,k,l]$ corresponding to the centroid of the detected SIFT feature $k,l$ in each sub-image $i,j$. This approach works in the presence of refractive features because, although textural blobs have distorted apparent motion in the \gls{LF}, their appearance is very similar across the \gls{LF} sub-images, particularly for small-baseline cameras like the hand-held Lytro employed in this work.

%\dorian{
% We rejected spurious \glspl{RLFF} based on three criteria: minimum number of sub-image matches, minimum diversity of views, and astigmatic lens model reprojection error. First, as described earlier, we rejcted features that did not match in at least $N_{MIN}$ sub-images.  We empirically selected $N_{MIN}$ to 12 out of 169 sub-images, to allow most putative features through. 

Next, to effectively estimate the interval of Sturm and extract the feature's parameters, we require that the sub-images observing the feature have sufficient diversity. If we consider the \gls{LF} as a grid of sub-images, a line of sub-images from this grid does not suffice to observe astigmatic refractions. Views must subtend a 2D space. To evaluate view diversity, we use the coefficient of determination $R^2$ of a line of best fit from the $s,t$ coordinates of matching views. High coefficients of determination correspond to a mostly linear set of views, and we empirically determined $R^2 > 0.65$ to be a suitable criterion for rejection. % Without this measure, 0-eigenvalues are produced from the eigendecomposition.  % which are problematic - no information on the second axis, makes the problem ambiguous
% any justification? low value: not enough samples to characterise the interval of Sturmm. Too high N_MIN, and it's too restrictive for our feature matching
% }

\subsection{Feature Extraction}

In this section we describe the process of estimating the parameters of the \gls{RLFF}. The feature detection yields a set of observations in discrete space $\vect{n_i}$. We begin by converting this to a continuous-domain representation by calibrating the camera to yield an LF intrinsic matrix~\cite{dansereau2013decoding}. This allows us to convert each observation $\vect{n_i}$ to a continuous-domain ray $\vect{\phi_i} = [s,t,u,v]$.

In the case of a Lambertian scene point, it is well established that the set of observations $\vect{\phi}$ will lie on a plane in the 4D light field~\cite{adelson1992single}. The geometry for this point-plane correspondence is depicted in Fig.~\ref{fig:lfDepthROVS}, and is given by
\begin{equation}
	\begin{bmatrix}	
 		u \\
  		v
	\end{bmatrix}	
	= \begin{pmatrix}
		-\frac{D}{P_{z}}
		\end{pmatrix}
 		\begin{bmatrix}	
			s - P_{x} \\
			t - P_{y}
 		\end{bmatrix}.
	\label{eq_Hyperfan_PtPlane}
\end{equation}
This can be interpreted as the intersection of two hyperplanes, where the slopes of the hyperplanes in the epipolar plane dimensions $s,u$ and $t,v$ are identical, given by $-D/P_z$.

We now generalise the point-plane correspondence by introducing an astigmatic lens between the camera and the Lambertian point, as depicted in Fig.~\ref{fig:rovsProjection}. We initially assume a thin toric lens with axes aligned with the $s$ and $t$ axes. This yields two orthogonal lines of focus at depths $P_{z1}, P_{z2}$. In $s,u$ the behaviour is similar to that of a Lambertian point at depth $P_{z1}$, whereas in $t,v$ the rays appear to emerge from a point at depth $P_{z2}$. We can write this more formally as
\begin{equation}
\begin{split}
	\begin{bmatrix}
    u \\ v
	\end{bmatrix}	
	=
	&S
	\begin{bmatrix}
    s - P_x \\ t - P_y
	\end{bmatrix}, \quad
	S =
	\begin{bmatrix}
	-D/P_{z1} & 0 \\ 0 & -D/P_{z2} 
	\end{bmatrix},
\end{split}
\label{eq_RLFF_GeneralizedPtPlane}
\end{equation}
where $P_z$ are apparent depths corresponding to the extremities of the interval of Sturm. 

For the more general case of non-camera-aligned lines of focus, i.e.~the case of a general astigmatic lens, we apply a transformation
\begin{equation}
\begin{split}
	\begin{bmatrix}
    u \\ v
	\end{bmatrix}	
	=
	&H
	\begin{bmatrix}
    s - P_x \\ t - P_y
	\end{bmatrix}, \quad
	H = V S V^\negative{1},
\end{split}
\label{eq_RLFF_LensAsEigenvectors}
\end{equation}
where $V$ is a rotation matrix for the special case of a rotated toric lenses, and the concatenation of two potentially non-orthogonal axes $[V_1,V_2]$ for a general astigmatic lens. This transformation assumes the interval of Sturm is close to parallel with the principal axis of the camera, an assumption that holds well under most of the imaging scenarios we consider, in which features must be visible through the \gls{RO}. 

Finally, separating the translation terms yields the generalised point-plane correspondence for points imaged through astigmatic lenses
\begin{equation}
\begin{split}
	\begin{bmatrix}
    u \\ v
	\end{bmatrix}	
	=
	&H
	\begin{bmatrix}
    s \\ t
	\end{bmatrix}
	+
	X
	, \quad
	X = -H 
	\begin{bmatrix}
	P_x \\ P_y
	\end{bmatrix}.
\end{split}
\label{eq_RLFF_Offsets}
\end{equation}
Note that this general form also describes the Lambertian case, for which the depths $P_{z1}$ and $P_{z1}$ are identical and \eqref{eq_RLFF_Offsets} degenerates to~\eqref{eq_Hyperfan_PtPlane}.

\subsection{Estimating Feature Parameters from Observations}
From~\eqref{eq_RLFF_Offsets}, observations of a scene point take the form
\begin{equation}
\begin{split}
	\begin{bmatrix}
    u \\ v
	\end{bmatrix}	
	=
	\begin{bmatrix}
    h1 & h2 & x1 \\ 
    h3 & h4 & x2
	\end{bmatrix}
	\begin{bmatrix}
    s \\ t \\ 1
	\end{bmatrix}.
\end{split}
\label{eq_RLFF_LinearSystem}
\end{equation}
Given a set of $[s,t,u,v]$ observations for a single feature, we find the least squares solution to \eqref{eq_RLFF_LinearSystem}, directly yielding estimates $\hat{H}$ and $\hat{X}$. 
% Dorian: I'm uncertain if the tilde notation is common in the literature to indicate estimation, ie \tilde{H} is an estimate of H.

Eigenvalue decomposition of $\tilde{H}$ allows us to estimate $V$ and $S$ % should be \tilde{V}, \tilde{S}?
following \eqref{eq_RLFF_LensAsEigenvectors}. Equating terms from \eqref{eq_RLFF_GeneralizedPtPlane}--\eqref{eq_RLFF_Offsets} allows us to solve for $P_x, P_y, P_{z1}, P_{z2}$ as well as the directions of the axes $V_1$ and $V_2$, $\theta_1$ and $\theta_2$, respectively. These are the parameters of the focal lines caused by an astigmatic lens and the interval of Sturm. These six parameters also compose our definition of the \textit{Refracted Light Field Feature (RLFF)}:
\begin{equation}
    RLFF = \begin{bmatrix} P_x, P_y, P_{z1}, P_{z2}, \theta_1, \theta_2 \end{bmatrix}.
    \label{eq_RLFF_deff}
\end{equation}

Though physically realizable $H$ matrices are symmetric, the estimate $\hat{H}$ may not be. Asymmetric matrices can result in imaginary eigenvalues. Prior to eigendecomposition we force $\hat{H}$ to be symmetric, $\hat{H}_S = (\hat{H}+\hat{H}^T) / 2$, where $^T$ is the matrix transpose. We 
%\begin{equation}
%\tilde{H}\sub{S} = \frac{\tilde{H} + \trans{\tilde{H}}}{2},
%\end{equation}
note that using a constrained least squares estimator would likely yield improved noise performance. 

When estimating the offsets, $P_x$ and $P_y$, we use the reconstructed $H\sub{R} = V S V^\negative{1}$ rather than $\hat{H}$, as this improves noise performance. The residual between $\hat{H}$ and $H\sub{R}$ forms a convenient indicator for outlier features not well described by our assumptions. 

\begin{figure} 
  \centering 
  \vspace{0.1em}
  \includegraphics[height=0.5\columnwidth]{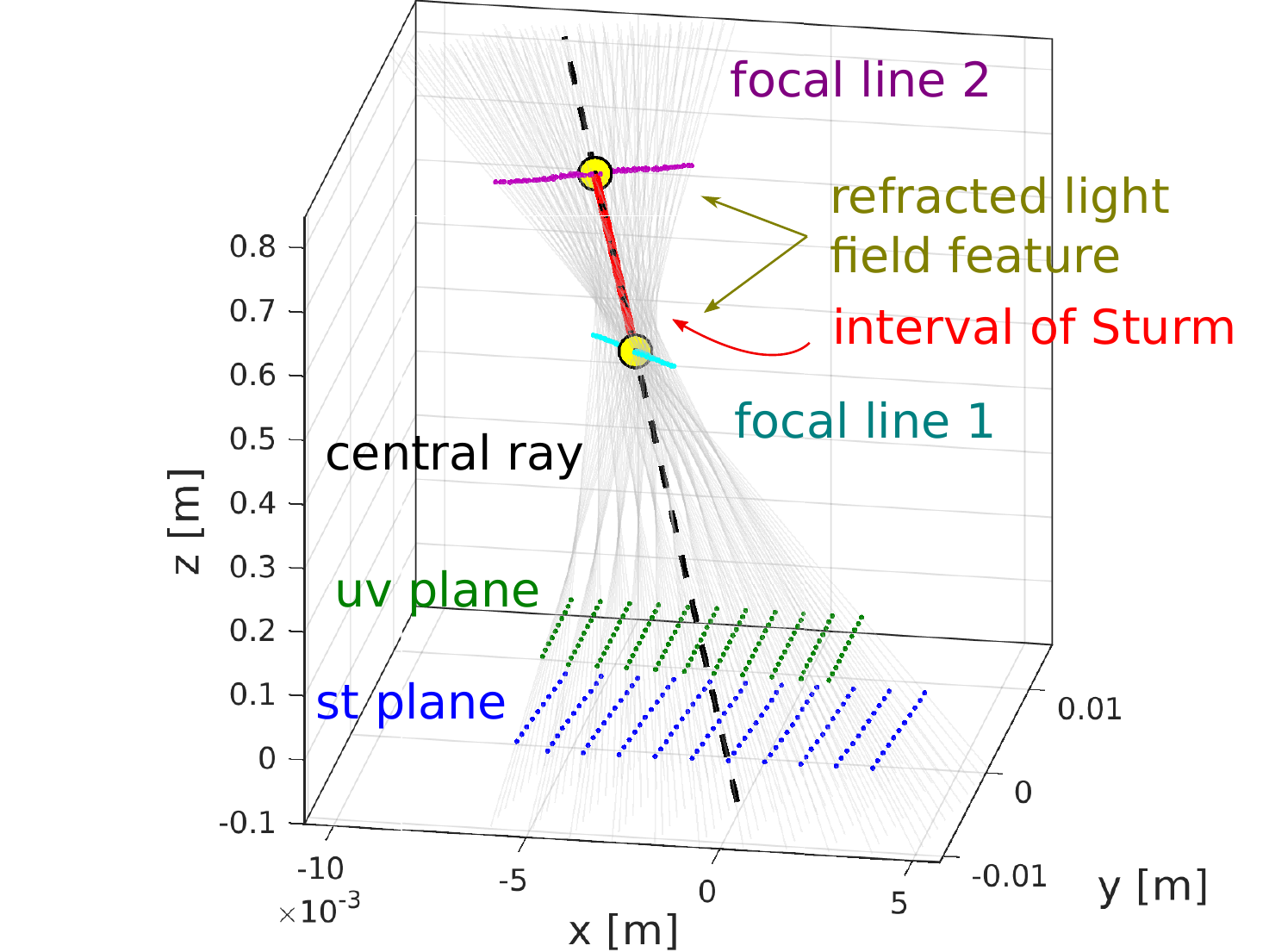}\label{fig:rlff_sample}
  \caption{An \gls{RLFF} extracted from Illum imagery. The $st$ sub-images (blue) and $uv$ observations (green) are shown from distance $D = 0.1m$ from the $st$-plane. The rays projected by $stuv$ (grey) pass through both the first and second focal lines (cyan and magenta, respectively). The central ray of the feature is shown (dashed black). The interval of Sturm (red), and the two 3D points (yellow) define our \gls{RLFF}.}
  \label{fig:3DlineSegment} 
\end{figure}

An example of a feature extracted from captured \gls{LF} imagery is shown in Fig.~\ref{fig:3DlineSegment}. The sub-image views are shown in blue, the set of $s,t,u,v$ observations across the \gls{LF} are shown in gray, and the estimated focal lines and interval of Sturm are shown as coloured line segments.

% For features that pass the first two tests, we applied the feature extraction process described in Section~\ref{sec:rlff_ext} to estimate the interval of Sturm. 

% Once the feature geometry is estimated, our final means of rejecting spurious detections is to test how well the astigmatic model fits. As noted earlier, the difference between estimated and reconstructed transformations $\tilde{H}, H_{R}$ is one indicator of how well the model describes the observations.  We also computed the reprojection error between measured and predicted $u,v$ observations in~\eqref{eq_RLFF_LensAsEigenvectors}. This allowed us to rejected features with high error in the higher-resolution sub-image space, again with a tunable threshold that we selected empirically to correspond to 0.25 pixels in the sampled $k,l$ space. % threshold set at 40 pixels, with no justification...???   % dd todo: 40 meters?!  40 pixels!?  both seem very high... I've guessed this means a total over all views of 40 pixels, i.e. 40 / 169 = 0.25 pix, but please confirm Dorian

\subsection{Driving SfM}
\label{sec_Driving_SfM}

The proposed feature definition comprises two infinite lines of focus and the interval between them. We anticipate constructing robotic vision algorithms that work directly off these features. However, existing systems like the popular SfM solution COLMAP~\cite{schoenberger2016sfmrevisited} accept only 2D image features like SIFT. Thus, we propose two mechanisms for driving existing vision systems with the proposed feature for comparison with monocular- and stereo-based SfM approaches. 

First, by projecting the 3D endpoints of the interval of Sturm into the central view of the camera, they are reduced to 2D image features. % that can be used as a drop-in replacement for existing features like SIFT.  
This \emph{RLFF mono} has the drawback of discarding all 3D information associated with the feature.  
Second, we therefore propose a variation of this approach in which we instead project the same endpoints into two separate LF views, separated by a baseline similar to that of the LF camera. This \emph{RLFF stereo} preserves most of the 3D information of the feature, discarding only the orientations of the focal lines. This also preserves our knowledge of the baseline over which depth is being estimated, an important detail for reconstruction algorithms that consider uncertainty associated with short-baseline depth estimates. We evaluate both variations of this approach in Section~\ref{sec:results}.

Note that our approach simultaneously detects both Lambertian and refracted features, and passes them all into \gls{SfM}. Lambertian scene points yield a zero-length interval of Sturm, and so appear as a single point rather than as a pair. This is visible in our results, e.g.~see the refracted and background Lambertian features in Fig.~\ref{fig:synth_stereo_rlff_ftr_motion}. 

Many applications will benefit by distinguishing Lambertian and refracted features, as Lambertian points generally correspond to an object's surface, while refracted features are images that exist in free space. Prior work has distinguished refracted features on the basis of slope differences in 3D subsets of the \gls{LF}~\cite{tsai2018refractedfeatures}. The \gls{RLFF} effectively measures how Lambertian an image feature is in a more complete way, as the entire \gls{LF} is employed. The interval of Sturm is of zero length for Lambertian scene points, and only takes on finite extent for refracted scene content.

Note also that in the case of refracted features we are passing two characteristic points to COLMAP as though they were separate 2D features. We require a descriptor for each of these to allow feature matching, but wish to disallow matching of front and back characteristic points $C_1$ and $C_2$ between frames. To this end we propose three approaches. First, the descriptor can be modified to reflect which focal line it belongs to, front or back, e.g.~through addition of a bias term, scaling factor, of by raising to some power. Then matches could not occur between front and back features as their descriptors differ substantially. Second, one can perform matching externally to the SfM tool while keeping track of which characteristic point each feature corresponds to. Then potential matches between front and back points are simply not evaluated. Finally, the approach taken in this work is to employ identical descriptors for the two points, yielding extraneous putative matches, and relying on outlier detection to reject these on the basis of epipolar geometry.

%!TEX root = main.tex

%\section{Experimental Results}

\section{Evaluation}
\label{sec:results}

\begin{figure}
\centering
\scriptsize
\setlength\tabcolsep{1.5pt} % default 6pt
\vspace{1em}
\begin{tabular}{ccc}
%\hline
	& SIFT & RLFF \\
%\hline
\raisebox{0.16\columnwidth}{\rotatebox[origin=c]{90}{Wine glass}}
	& \includegraphics[width=0.47\columnwidth,height=0.34\columnwidth]{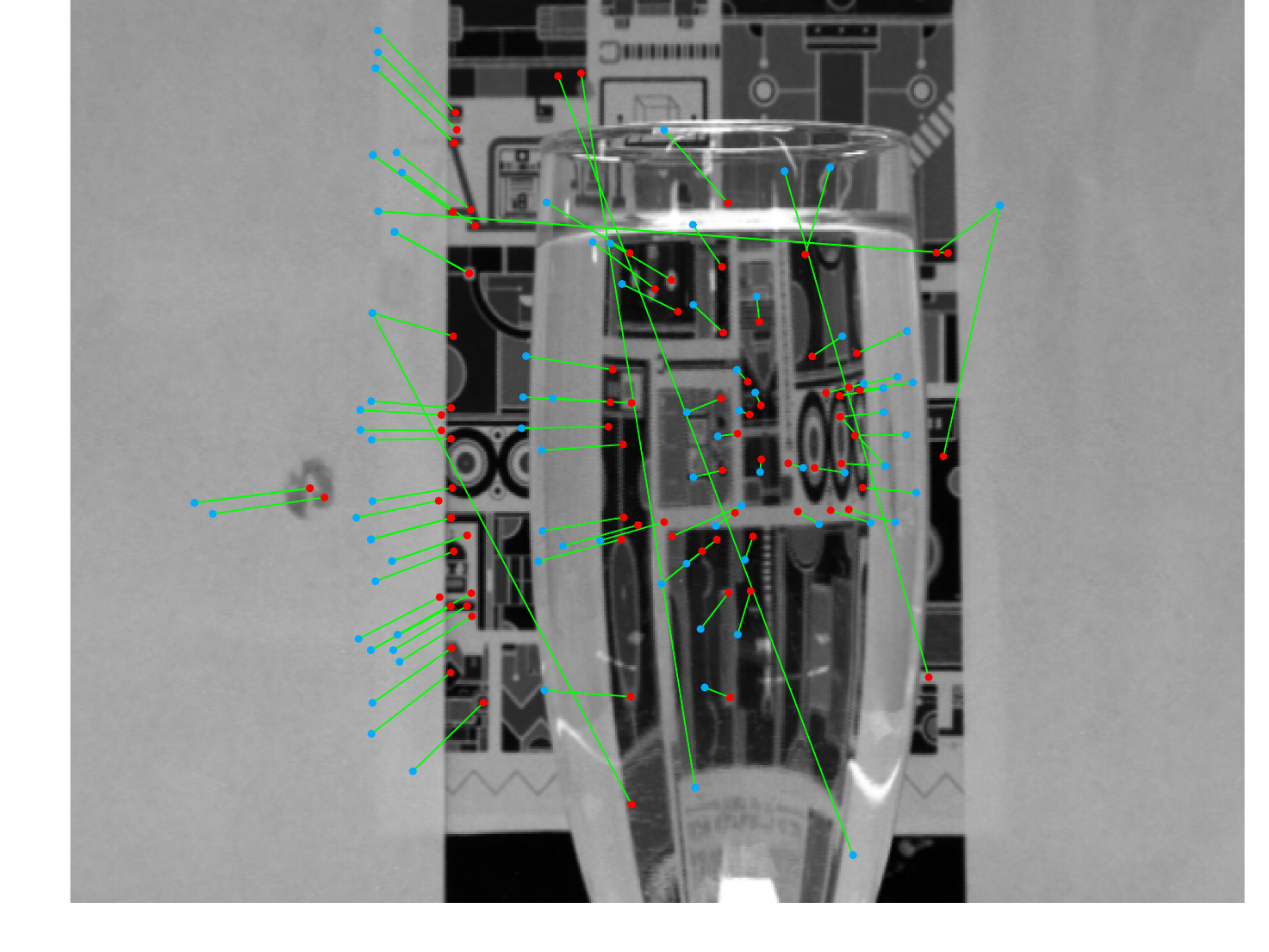}\hfill
	& \includegraphics[width=0.47\columnwidth,height=0.34\columnwidth]{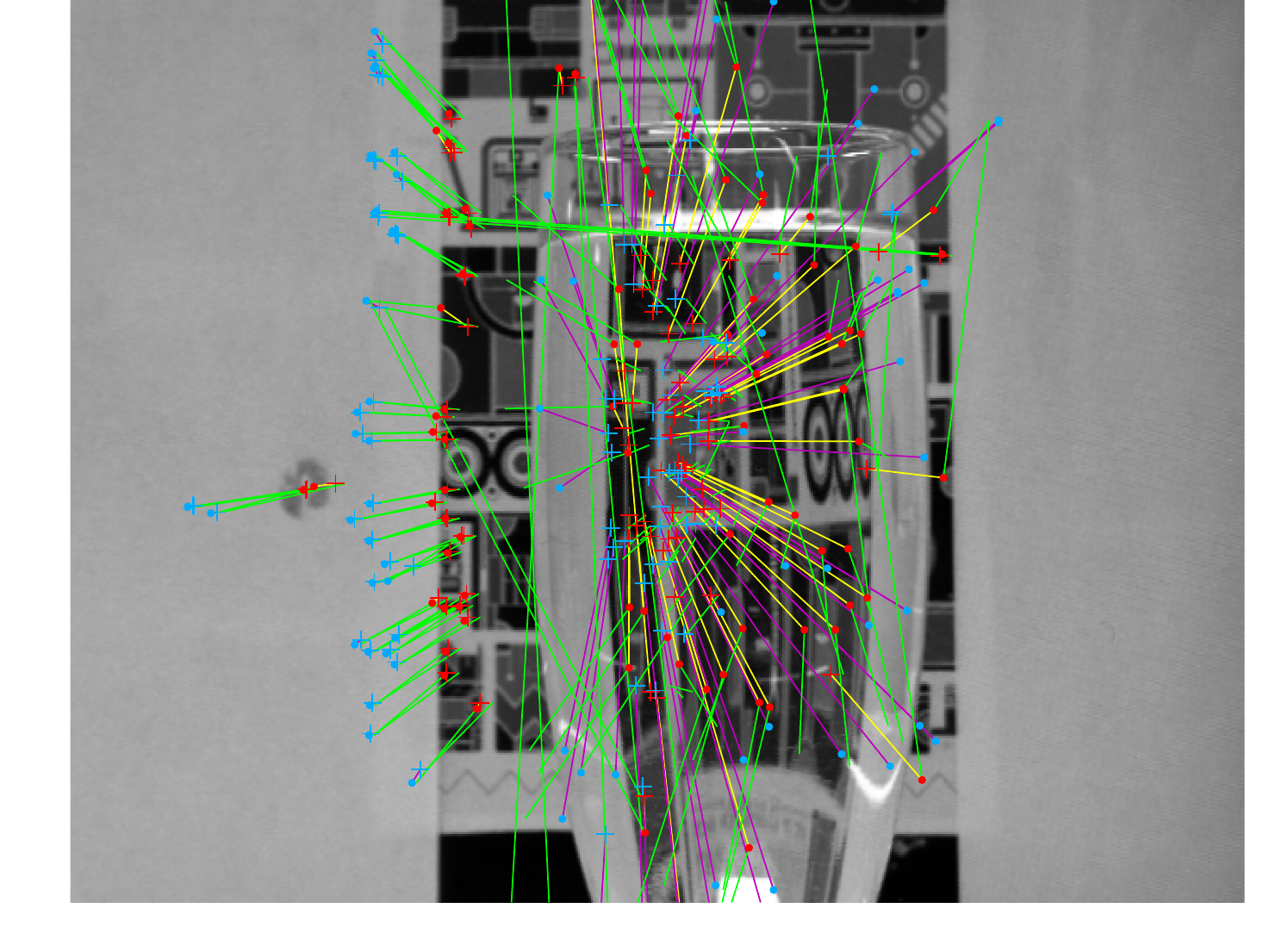}
	\\
\raisebox{0.16\columnwidth}{\rotatebox[origin=c]{90}{Sphere}}
	& \includegraphics[width=0.47\columnwidth]{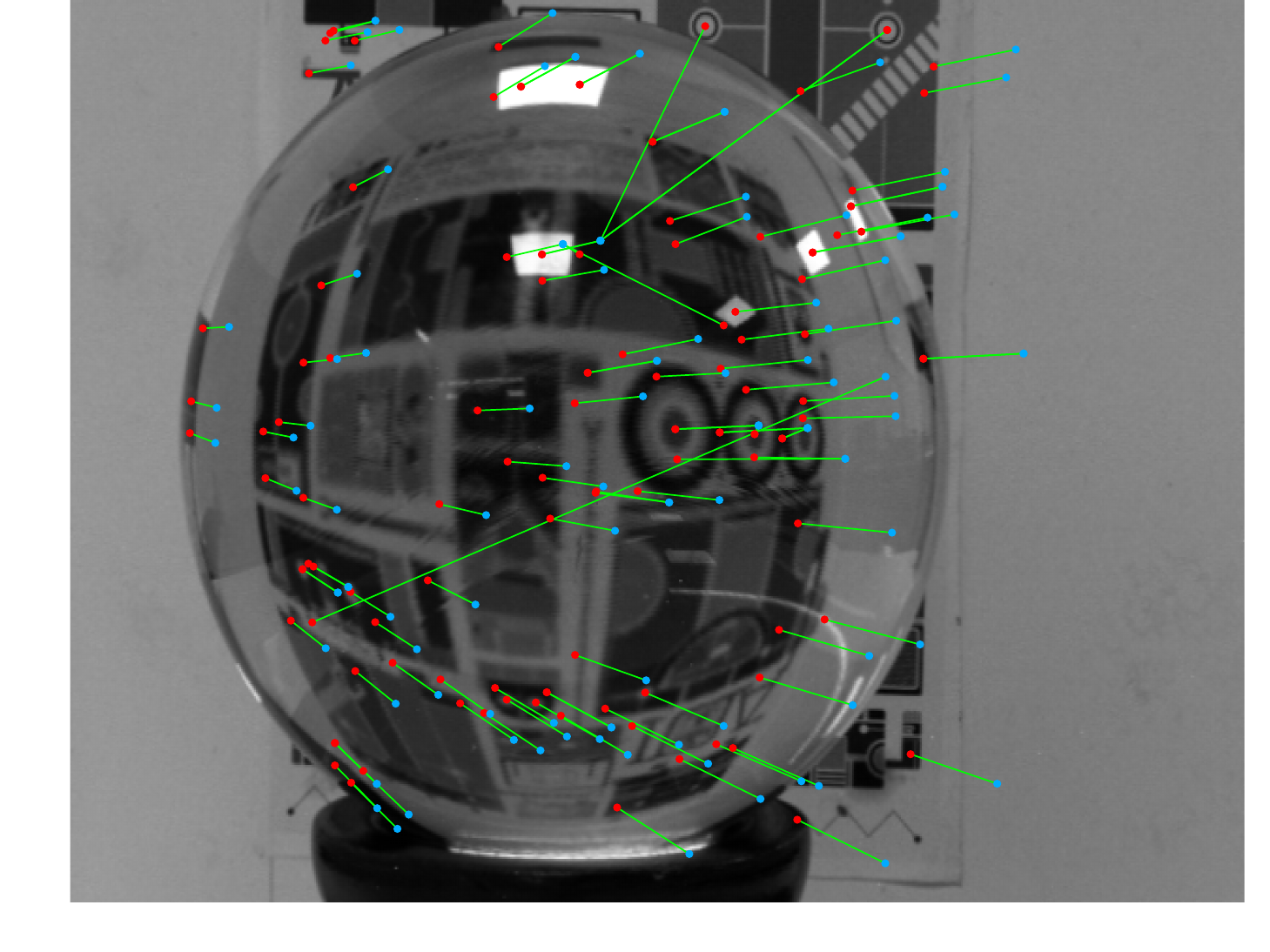}\hfill
	& \includegraphics[width=0.47\columnwidth]{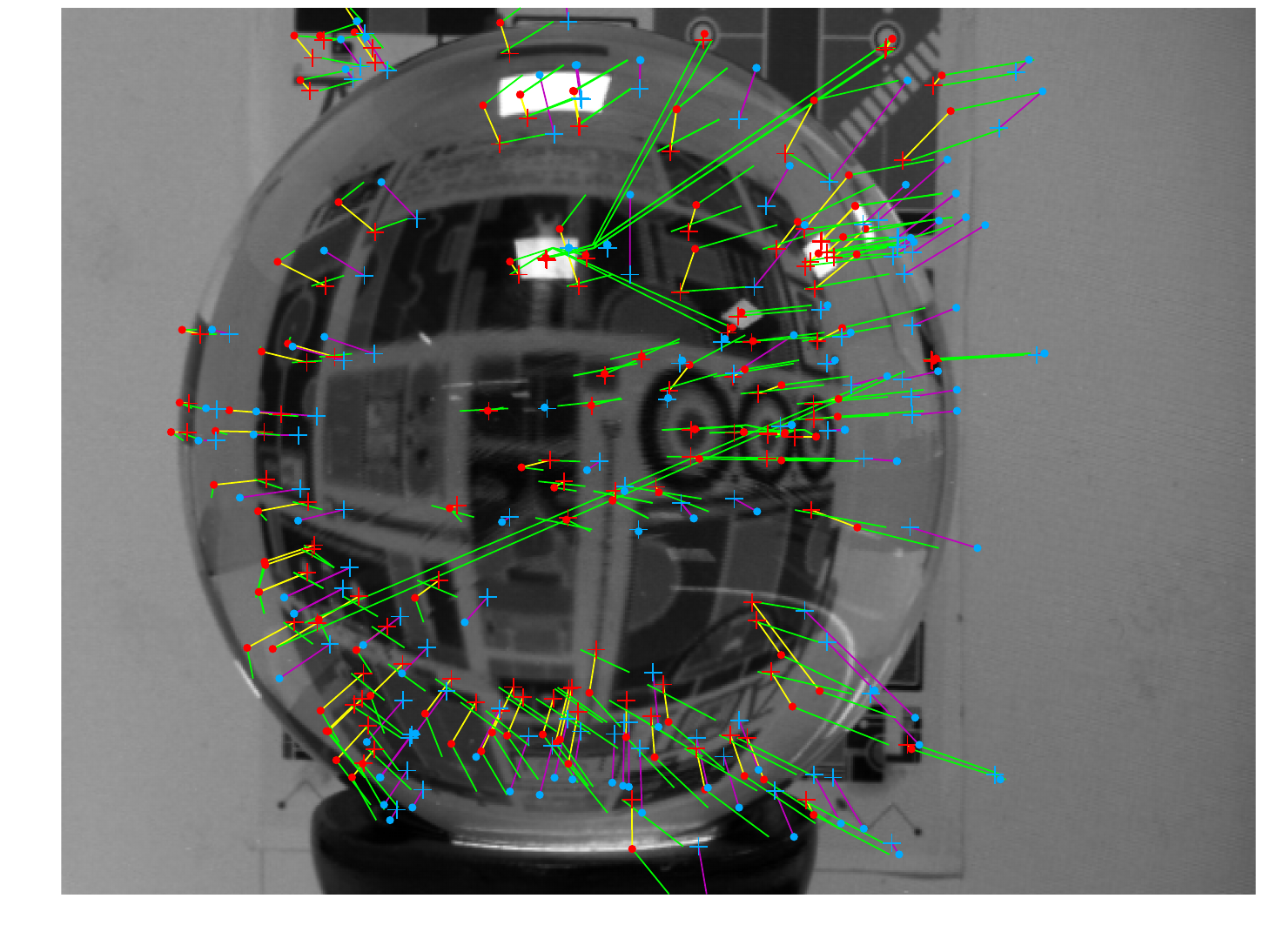}
	\\
\raisebox{0.16\columnwidth}{\rotatebox[origin=c]{90}{Cylinder}}
	& \includegraphics[width=0.47\columnwidth]{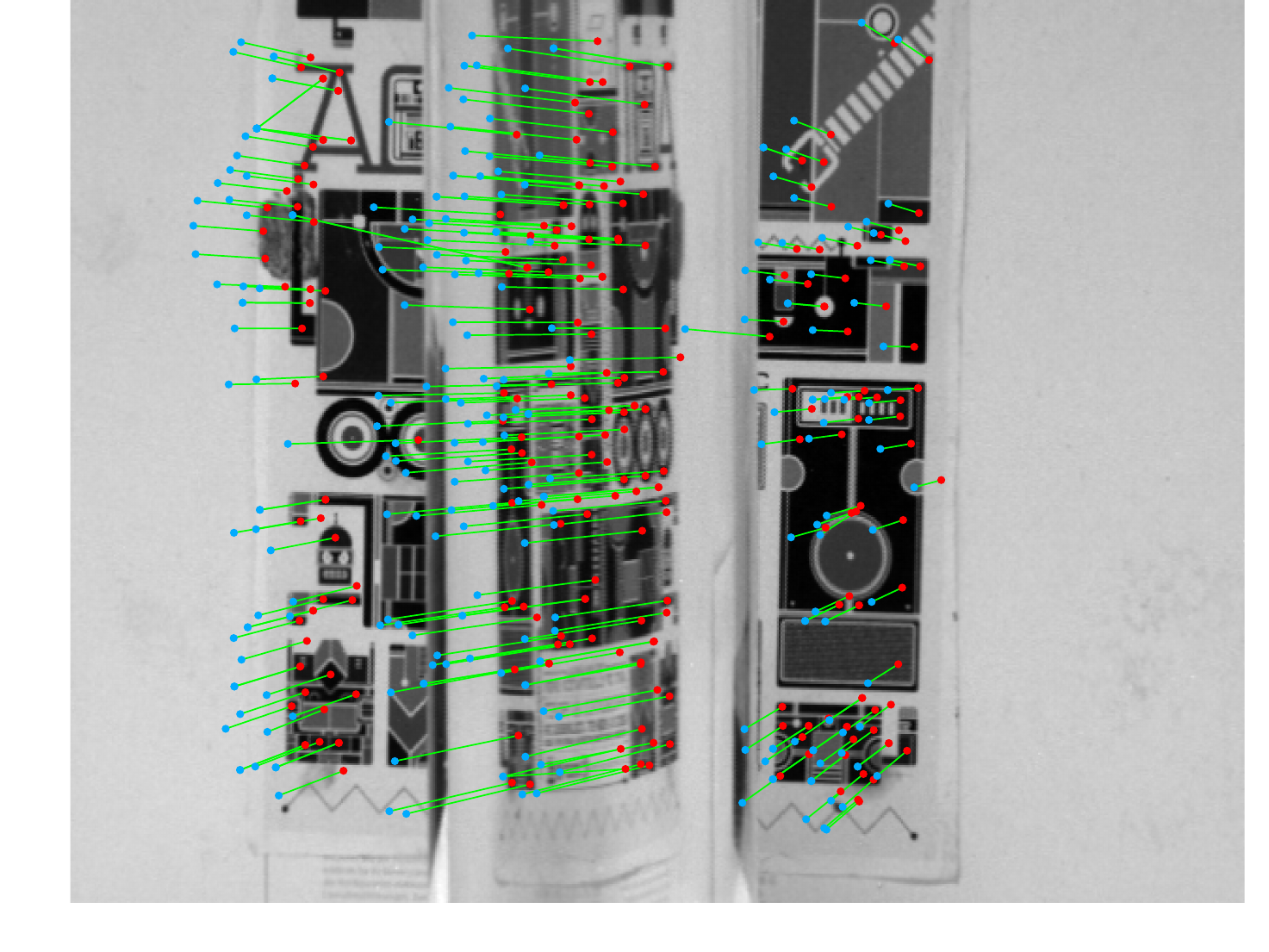}\hfill
	& \includegraphics[width=0.47\columnwidth]{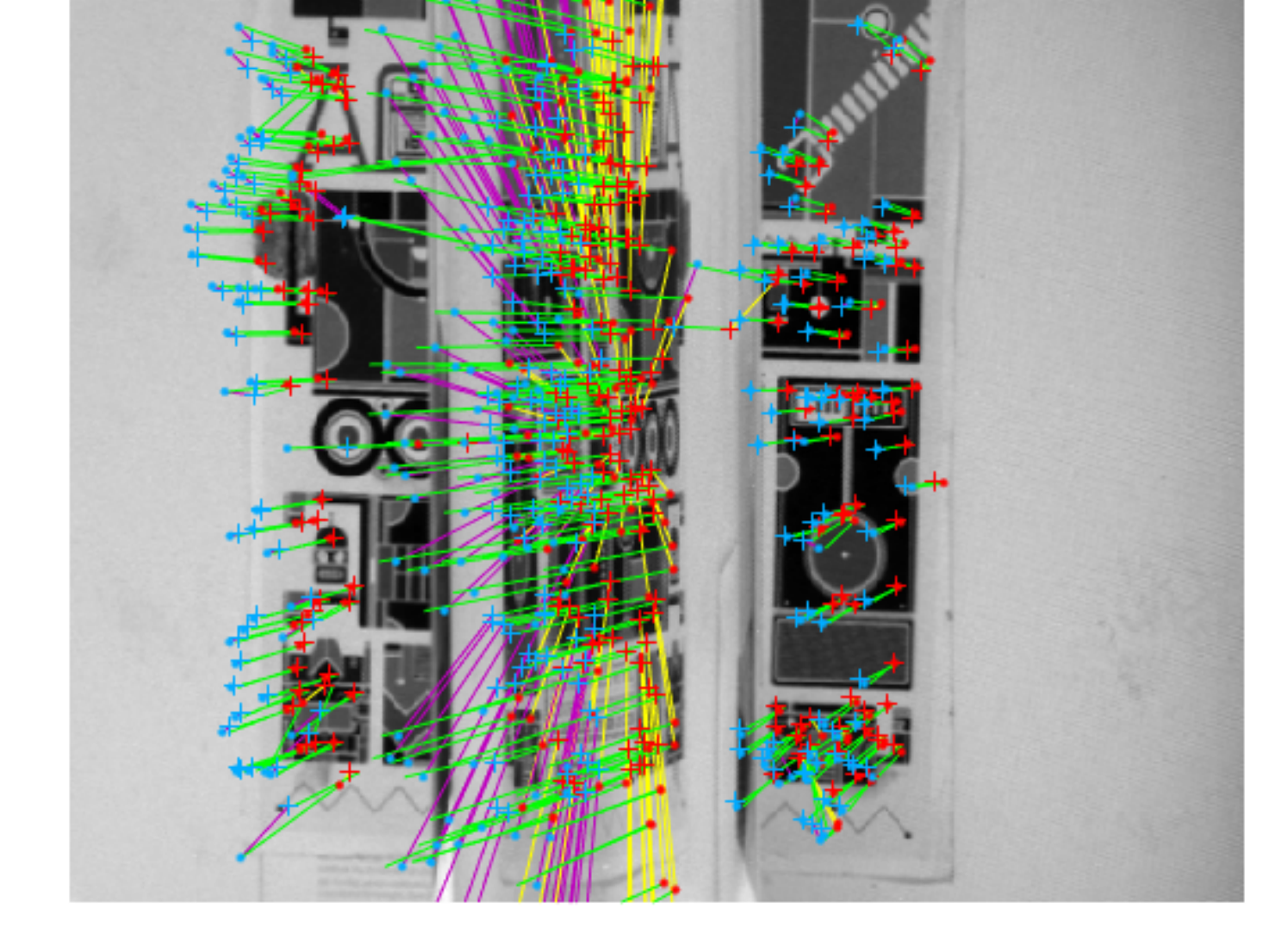}
	\\
\end{tabular}
\caption{Three example scenes for which COLMAP fails when using SIFT features (left) but succeeds when using the proposed \gls{RLFF}. These examples also show different types of camera motion: the wine glass has forwards motion, the sphere has horizontal motion, and the cylinder has both. The proposed features show some inconsistent apparent motion near the edges of the refractive objects; however their motion is more consistent than SIFT, allowing COLMAP to converge.
}
\vspace{-1.0em}
\label{fig:siftfail}
\end{figure}

To evaluate the proposed feature we mounted an Illum \gls{LF} camera on a Franka Emika Panda robot arm. The \gls{LF} camera was calibrated using the \gls{LF} Toolbox~\cite{dansereau2013decoding}. To reduce the effect of extreme lens distortion, we cropped the $15\times15$ \gls{LF} to a $13\times13$ array of sub-images. The robot arm was used for repeatability and ground truth trajectories with sub-millimeter positional accuracy. We used a variety of \glspl{RO}, such as a water bottle, glass sphere, glass cups filled with water, combined with a variety of textured backgrounds. % \dorian{TODO, reference figure of different objects/backgrounds?}. 
The experimental setup is illustrated in Fig.~\ref{fig:robotview}.

\subsection{Structure from Motion}

%Although not investigated in this paper, we can also calculate slope inconsistency ratio, $c = \sqrt{(\lambda_1 - \lambda_2)^2/(\lambda_1 + \lambda_2)^2}$. This measure can be used to determine if the feature is refracted or not. As in~\cite{tsai2018refractedfeatures}, the non-refracted features can be used to identify Lambertian scene content. We also anticipate that finding $c$ is more robust than previous approaches~\cite{tsai2018refractedfeatures}, as we now use the entire \gls{LF}, rather than only the central vertical and horizontal sub-images.  % however, we provide no evidence in support of this claim!
%Finally, for inter-frame matching (between \glspl{LF}), we only match centre-view image features using Euclidean distance. In the future work, we will consider 4D manifold-based feature matching, as in~\cite{dansereau2019liff} and~\cite{zhang2017ray}.

% incredibly rough first pass!
To evaluate the \gls{RLFF}, we used the popular COLMAP \gls{SfM} implementation~\cite{schoenberger2016sfmrevisited}. Following the procedure outlined in Section~\ref{sec_Driving_SfM}, we converted each detected \gls{RLFF} to a pair of 2D image features, as seen from the central view of the LF camera (the middle view of a LF if we consider the LF as a grid of views). We also evaluated the alternative approach of projecting the feature into a stereo pair of LF views, preserving depth information. 

For comparison, we evaluated conventional SIFT features as seen in the central LF view. To better understand the impact of projecting features into stereo pairs, we also compared with SIFT similarly projected into a pair of LF views, with the same baseline as for the \gls{RLFF} test.  In all, we compared four methods using COLMAP, covering the proposed \gls{RLFF} and SIFT for both monocular and stereo views: RLFF mono, RLFF stereo, SIFT mono and SIFT stereo.
% the proposed \gls{RLFF} feature and SIFT, passed to COLMAP via a single monocular view, or via a stereo pair of views.

% do we need to explain percent pass for convergence?
We evaluated performance using COLMAP's sparse SfM. SfM was not able to reconstruct all scenes for all types of objects, as its solution did not always converge. This was especially problematic for scenes dominated by refracted content. Following~\cite{dansereau2019liff} and~\cite{schoenberger2016sfmrevisited}, we evaluated peformance in terms of the percentage of scenes for which COLMAP converged, the number of image features per image, putative image feature matches per image, inlier matches per image during \gls{SfM}, putative match ratio, mean number of 3D points, track length, precision and matching score. The \textit{putative match ratio} is the proportion of detected image features that yielded putative matches.
The \textit{mean number of 3D points} in the reconstructed models serves as an indicator of how many features were stable enough to be included into the model.
The \textit{track length} is the mean number of camera poses over which a feature was successfully tracked. 
The \textit{precision} is the proportion of putative image feature matches that yielded inlier matches.
The \textit{matching score} is the proportion of image features that yielded inlier matches. 
Note that for the \gls{RLFF} features, we divided the number of image features, putative matches, inlier matches and 3D points by two because a single \gls{RLFF} is represented by a pair of points, the extents of the interval of Sturm. Similarly, we divided the track length by two for both stereo approaches, as twice the number of images were considered during the motion sequences. This yielded more meaningful quantitative comparison.

We collected 20 sequences consisting of 10 to 20 camera poses each, covering a variety of motion trajectories and a diverse set of scene content including spherical, cylindrical, and general astigmatic elements. The dataset contains 218 \glspl{LF} in total. Example images highlighting the differences between SIFT and \gls{RLFF} performance are shown in Fig.~\ref{fig:siftfail}. In all of these scenes COLMAP was unable to converge while using either version of SIFT, but succeeded when using our \gls{RLFF}.

\begin{figure}
 \centering
 \vspace{1em}
 %\subfloat[][]{\includegraphics[,width=0.9\columnwidth]{Figures/IMG_6699__Decoded_slope_abs_diff.png}
 % \label{fig:rlff_slope_diff}}\\
 \includegraphics[width=0.75\columnwidth]{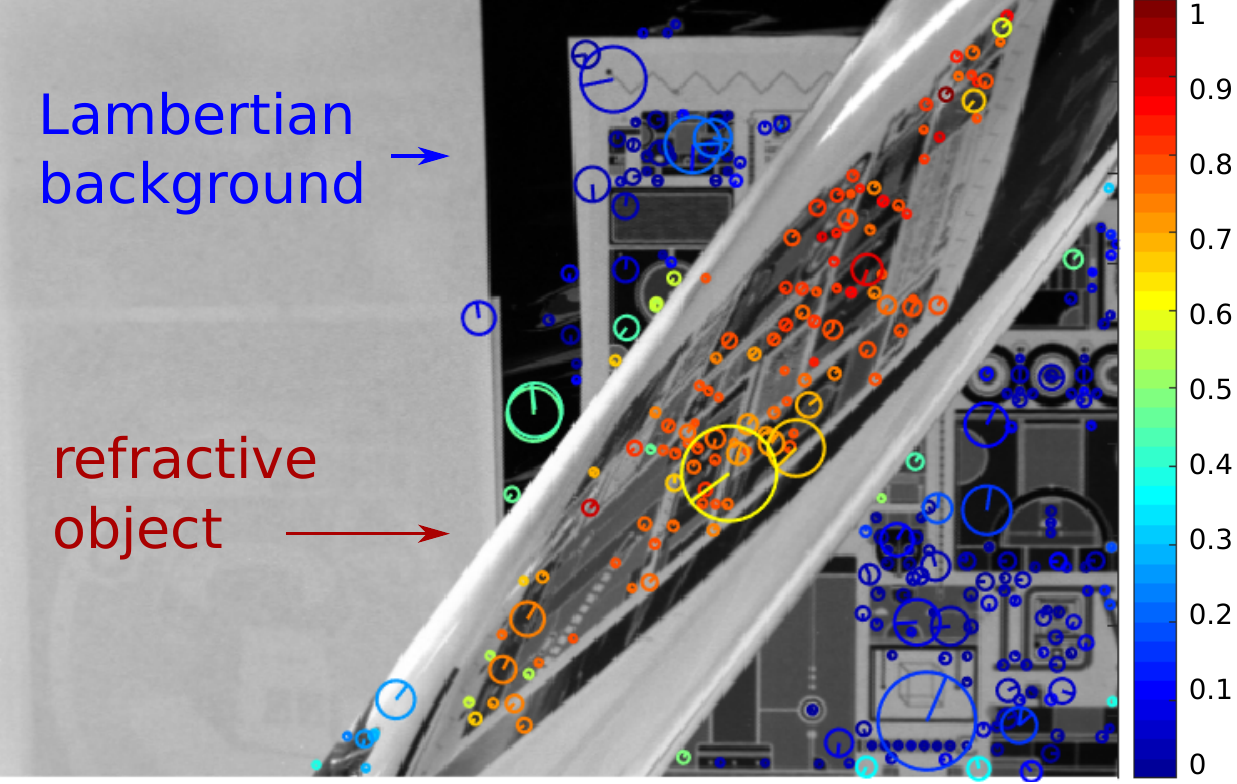}
 %\includegraphics[width=0.8\columnwidth]{Figures/IMG_6699__Decoded_slope_abs_diff_cropped.png}
  %\subfloat[][]{\includegraphics[,width=0.475\columnwidth]{Figures/20200619-1440_rlff_cyl_diag_rlff_intervalsOfSturmm.pdf}
  %\label{fig:rlff_intervalOfSturmPoints}}\hfill
  %\subfloat[][]{\includegraphics[width=0.475\columnwidth]{Figures/20200619-1440_rlff_cyl_diag_synthStereo_sfm_reconstruction.pdf} 
%\label{fig:synth_stereo_reconstruction}}
   \caption{
    %For the scene of a refractive cylinder (a), a side-by-side illustration of the (b) pair of 3D points representing the intervals of Sturmm for every \gls{RLFF} in the image, and (c) the reconstruction of scene from synthetic stereo \gls{RLFF}. Both 3D scenes closely resemble the scene geometry in that a a large diagonal section of features extends where the refractive cylinder is observed.
    Identifying features as Lambertian or refracted. Features are coloured to indicate the extent to which they have consistent slopes following \eqref{eq_Hyperfan_PtPlane}. Warmer colours indicate very inconsistent slopes and thus refracted features. Some inconsistencies are observed along occlusion boundaries and image borders (green/turquoise features), due to the apparent slope inconsistencies these induce in the \gls{LF}.
   }
   \label{fig:rlff_3D_reconstruction}
\end{figure}

\begin{figure}
  \centering
  \includegraphics[width=0.95\columnwidth]{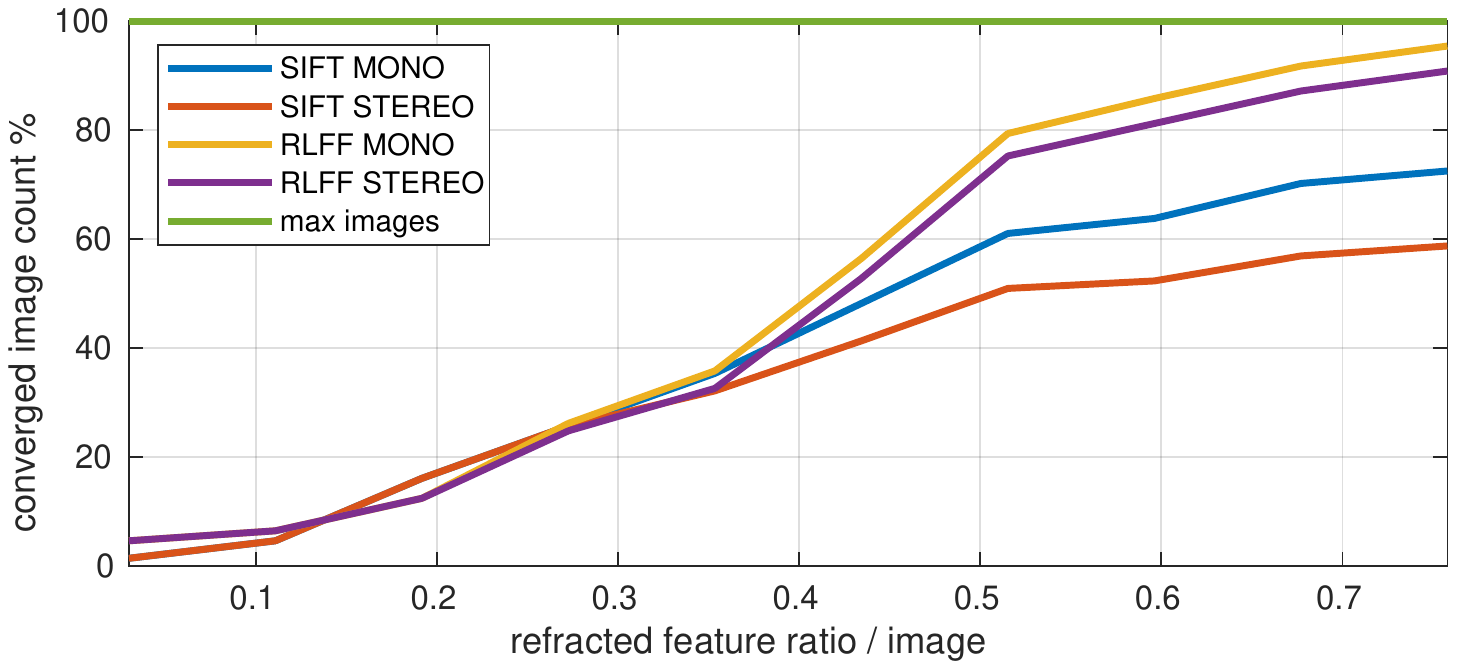}
  \caption{Cumulative histogram of imagery successfully incorporated into the COLMAP SfM model (as a percent of 218 LFs) versus the refracted scene content in the image. Toward the right of the figure, imagery becomes increasingly challenging as more of the image features become refracted and fewer images are being incorporated by each SfM solution, especially for the SIFT-based solutions. An ideal method would incorporate all images, ending with a value of 100 \%, being able to incorporate all of the imagery. Our RLFF-based SfM reconstructs more images in total, showing the strongest advantage with more refracted content.
  %The proposed method reconstructs more images in total, showing the strongest advantage with more refracted content.
  % \dorian{TODO: guide the reader, explain what it is, what an ideal method should look like, how SIFT-based approach is failing and how/where ours is succeeding. Define refracted feature ratio in text.}
  }
  \label{fig:cumhist}
\end{figure}

\begin{ctable}[%
 caption = {Evaluating the proposed \gls{RLFF} and SIFT in SfM, both in monocular and stereo modes. More scenes converge using the proposed method, and it outperforms the state-of-the-art in all measures except precision.},
 label = tbl:sfmFeatureStats,
 doinside = \scriptsize,
 %width = \textwidth,
%  pos = t!,
 star
 ]
 {lccccccccccc}{
 }
 {\FL Methods & \# \gls{LF}s & {\%}  & {\# \gls{LF}s} & {Image} & {Putative} & {Inlier} & {Putative} & {3D} & {Track} & {Precision} & {Matching} \\
  {} & {converged} & {Pass} & {common} & {Features} & {Matches} & {Matches} & {Match} & {Points}  & {Length} & {} & {Score} \\
 & & & & /\ Img & /\ Img & /\ Img & Ratio & & & &   \ML
    SIFT MONO &  158 & 0.75 & 118 &    372 &    122   &    112    &   0.369   &   431  &   5.11    &   0.916    &  0.340  \\
    SIFT STEREO & 128 & 0.6 & 118 &    370  &    131    &   122   &    0.369  &   524   &   4.16    &  \textbf{0.922}  &    0.342  \\
    \gls{RLFF} MONO &  208 & \textbf{0.95} &  118 &       321 &     155   &     134  &    0.448   & 437  &   \textbf{5.27}  &    0.842   &   0.385   \\
    \gls{RLFF} STEREO & 198 & 0.9 &  118 &       321  &    \textbf{168}    &   \textbf{147}   &    \textbf{ 0.487 }&  \textbf{684}  &   4.04  &    0.863   &  \textbf{0.426} \LL
 }
\end{ctable}

\begin{figure}
 \centering
  \subfloat[][]{\includegraphics[,width=0.9\columnwidth]{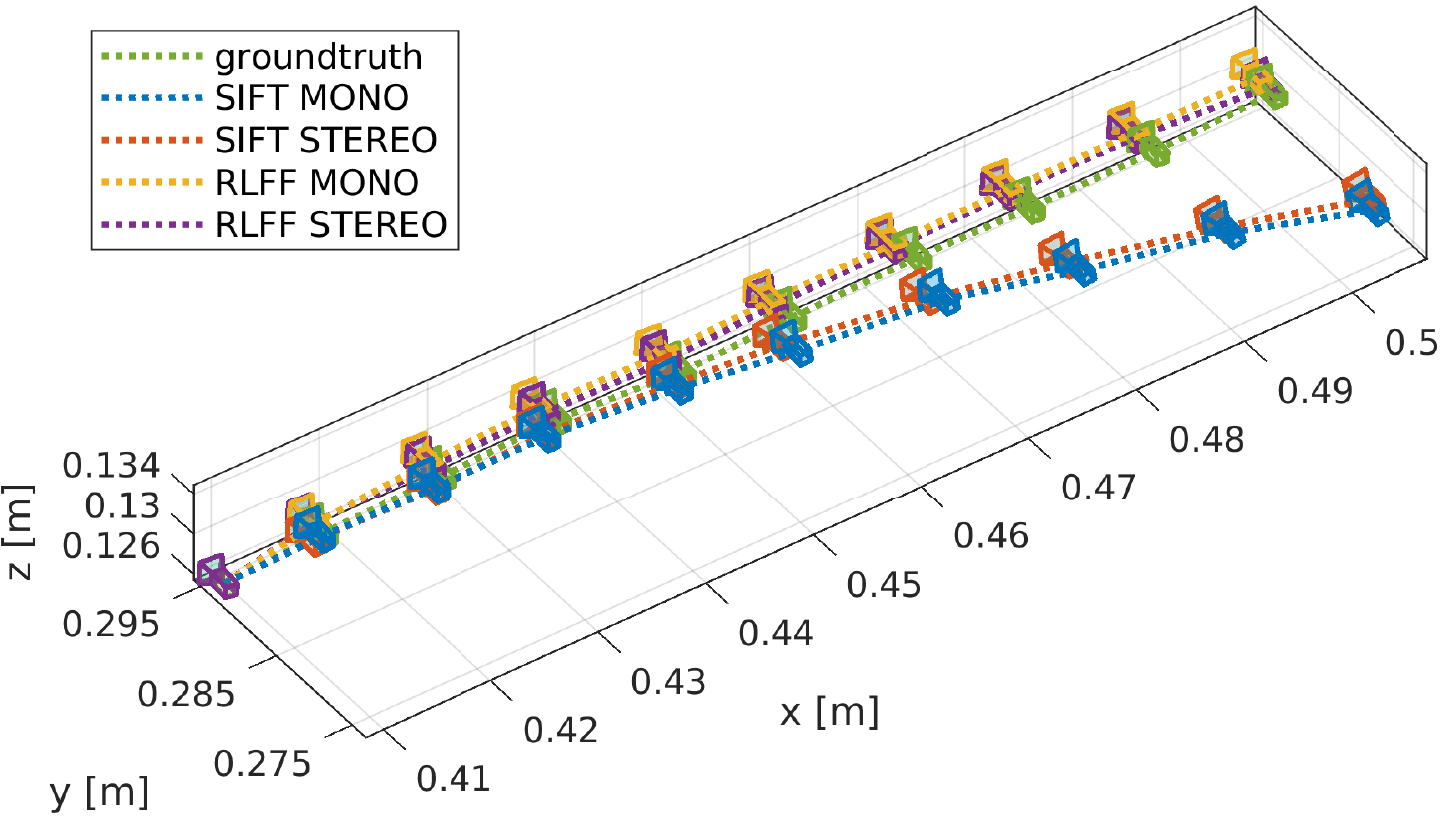}
  \label{fig:pose_trajectories}}\\
  \subfloat[][]{\includegraphics[width=0.48\columnwidth]{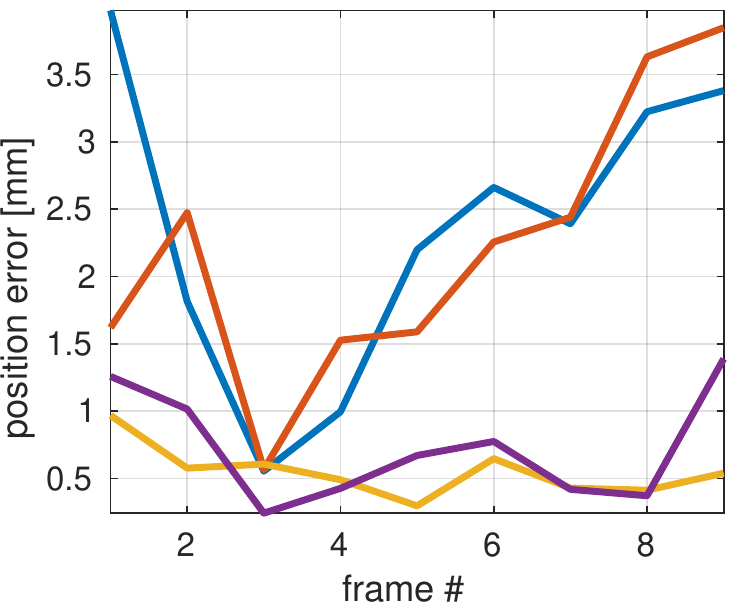} 
\label{fig:incremental_position_error}}\hfil
\subfloat[][]{\includegraphics[width=0.48\columnwidth]{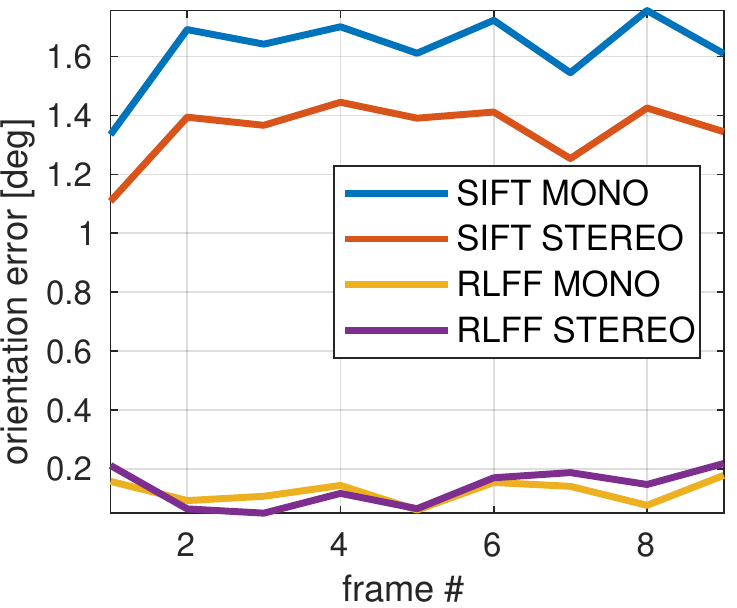} 
\label{fig:incremental_orientation_error}}
   \caption{Comparing pose accuracy in a trajectory that approaches a refractive object, making images progressively more difficult due to increasing refractive scene content: (a) Estimated trajectories show both proposed stereo and mono variants of \gls{RLFF} outperforming both variants of SIFT; (b) Relative, instantaneous translational and (c) rotational error show the proposed methods outperforming SIFT in all cases. 
   }
   \label{fig:rlff_glass_towards_object}
\end{figure}

To understand how \glspl{RO} affect SfM performance, we evaluated the extent to which each LF is dominated by refractive features.  We took the extent of the interval of Sturm for each feature as an indication of how Lambertian it is. Lambertian scene points evaluated by our feature extractor yield an interval of Sturm of zero length, or equivalently, correspond to a plane with equal slopes in horizontal and vertical dimensions following~\eqref{eq_Hyperfan_PtPlane}. Fig.~\ref{fig:rlff_3D_reconstruction} shows an example.

A limitation of this approach is that it does not work for spherical lenses, which produce a well-formed image that behaves identically to Lambertian scene content. Human viewers are also susceptible to this, and it is the basis for some display technologies that produce images floating in air. Features refracted through spherical objects are thus not well detected by our method. They do, however, make excellent points for use in SfM, so while we do not detect them as being refracted, we do make use of them in the SfM solution.  %Another limitation of this approach is that it depends on strong camera calibration and rectification, without which all Lambertian features appear as though they were refracted content. 

% if the sub-images are inconsistent with Lambertian feature motion. This was observed in the ``corners'' of the sub-images within the \gls{LF}. As stated earlier, we cropped the sub-images to a $9\times9$ image array to reduce this effect.  % pretty certain I state something similar earlier in setup/implementation

We plotted the number of images correctly incorporated into an SfM solution by COLMAP. By sorting images on the horizontal axis according to the fraction of features identified as refractive (i.e. the refracted feature ratio), we obtain the cumulative histogram shown in Fig.~\ref{fig:cumhist}. This shows both variants of the proposed method enabled SfM to succeed with almost all images, while the SIFT-based methods failed to converge for a significant number. Importantly, the difference in performance is due chiefly to scenes dominated by refracted content.

\subsection{3D Reconstruction Performance}  

We ran COLMAP across all of the collected scenes, and summarise the results in Table~\ref{tbl:sfmFeatureStats}. The statistics show that \gls{RLFF} methods have a higher proportion of putative and inlier matches, the number of 3D points in the reconstructions, the track length, and the overall matching score. Our methods converged 15-35\% more frequently than their SIFT counterparts, and performed better in almost all measures except precision. The stereo version of \gls{RLFF} generally showed higher performance, though the monocular version allowed more scenes to converge. 
% do we have a good explanation for mono over stereo? Perhaps just that Colmap is optimised for mono, so our stereo work-around was just to throw in the stereo images in sequence? The baseline is not really exploited by the SfM framework. Wish there was an easy-to-use stereo SfM framework.
%The precision result indicates that the SIFT-based approaches had more putative matches that were used as inliers in the final \gls{SfM} solution. 
We explain weaker precision performance of \gls{RLFF} by noting more putative matches due to doubling the number of 2D image features, which subsequently quadrupled the number of potential 2D feature matches. %projecting $c_1$ and $c_2$ into the sub-images. This quadrupled the potential 2D feature matches. 
In this preliminary work, descriptors were only duplicated for the \gls{RLFF} projections to 2D image features. For refracted image features, the distance between feature projections was sufficiently large to discriminate. However, for Lambertian features where the projections are coincident, we increased the putative matches without increasing the inliers. This resulted in lower precision. To prevent this issue, we can adopt the strategies proposed earlier in Sec.~\ref{sec_Driving_SfM} for future work. Overall, these results showed that the proposed \gls{RLFF} feature allowed COLMAP to operate in many scenes for which it could not previously operate.

\subsection{Camera Trajectory Estimation}

We also evaluated the accuracy of camera trajectories estimated by SfM. Ground truth was available by virtue of our use of a robotic arm to carry out the camera trajectories. A comparison of camera trajectory estimates and their corresponding translational and rotation error are shown in Fig.~\ref{fig:rlff_glass_towards_object}. For this scene, the proposed \gls{RLFF} feature showed substantially higher performance in both rotational and translational error, yielding a much more accurate trajectory estimate. 

\begingroup
\setlength{\tabcolsep}{3.0pt}
\begin{ctable}[%
 caption = {Comparison of relative, instantaneous pose error using monocular and stereo \gls{RLFF} and SIFT. The proposed \gls{RLFF} allowed almost all sequences to converge, while SIFT did not. The proposed feature outperformed SIFT both on the easier sequences in which all methods converged (top), and in the more challenging sequences (bottom).},
 label = {tbl:SfMcomparison},
 doinside = \scriptsize
%  pos = t!
 ]
 {cc cc  cc  cc  cc}{ 
%^\tnote[a]{\dorian{Can add seq when unfilt (FAIL), but filt succeeded (ipr)}}
 }
 { \FL \multicolumn{2}{c}{\textbf{}}  & \multicolumn{2}{c}{{SIFT}} & \multicolumn{2}{c}{{SIFT}} & \multicolumn{2}{c}{{RLFF}} & \multicolumn{2}{c}{{RLFF}} \\
 & & \multicolumn{2}{c}{{MONO}} & \multicolumn{2}{c}{{STEREO}} & \multicolumn{2}{c}{{MONO}} & \multicolumn{2}{c}{{STEREO}} \\
 {Seq.} & {\#LFs} & {$e_{tr}$}~[mm] & {$e_{rot}$}~[deg] & {$e_{tr}$} & {$e_{rot}$} & {$e_{tr}$} & {$e_{rot}$} & {$e_{tr}$} & {$e_{rot}$} \ML
 %    Trial    #LF      et_sm        eo_sm        et_ss        eo_ss        et_rm         eo_rm        et_rs         et_ro  
 %   _____    ___    _________    _________    _________    _________    __________    _________    __________    _________
%
      1 &      10  &  1.37  &  0.40   & 1.34   &  0.40   &  2.64  &  0.20    & 1.03  &  0.19 \\
     2   &   10    &  1.22  &  0.27   & 1.44   & 0.29    & 2.36   & 0.28     &1.58   & 0.21 \\
     3   &   10    &  16.50 &     6.86&    2.08&    0.11 &    5.83&     0.29 &    2.51&    0.23 \\
     4   &   10    &  1.99  &  0.14   &  2.12  &  0.13   &  1.26  &  0.13    & 2.17   & 0.15 \\
      5   &   10    &  2.35  &   1.62  &  2.22  &    1.35 &   0.55 &   0.12   & 0.73   & 0.14 \\
      6   &   10    &  0.84  &  0.47   & 2.97   & 0.42    &0.93    &0.42      & 3.07   & 0.43 \\
      7   &   10    &  1.60  &   1.52  &  4.02  &   2.65  &  0.92  &  0.17    & 2.38   &  1.78 \\
     8   &   10    &  4.21  &  0.60   & 4.69   & 0.53    & 7.20   & 0.71     &3.63   & 0.38 \\
     9   &   10    &  5.82  &   1.53  &  5.71  &   1.26  &   5.43 &   0.78   &  15.49 &    2.88 \\
     10   &   18    &  2.21  &  0.97   & 6.97   &  1.91   &  1.42  &  0.78    &  9.47  &   3.34 \\
     11   &   10    &  24.44 &  8.51   & 24.38  &   7.38  &    7.87&     0.69 &    1.98 &   0.34 \ML
Mean  &   -  &  5.69   &   2.08  &  5.27   &  1.49 &   \textbf{3.31}  &  \textbf{0.42} &    4.00  &  0.92 \ML
      12  &    10   &    7.24 &    2.85 &   7.90   &  3.01  &   1.87  &   1.33   &     -  &      - \\
      13  &    10   &   1.54  &  0.26   &    -     &   -    &    -    &    -    &    -    &    - \\
      14  &    10   &   7.13  &   1.23  &     -    &    -   &  14.70  &  0.48  &   1.87   & 0.16 \\
      15  &    10   &   2.30  &  0.62   &    -     &   -    & 1.75    &0.36    & 2.65    &0.50 \\
      16  &    10   &      -  &      -  &     -    &    -   &  1.19   & 0.45   &  1.31   & 0.34 \\
     17  &    20   &      -  &      -  &     -    &    -   &  1.39   & 0.19   &  1.15   & 0.15 \\
     18  &    10   &      -  &      -  &     -    &    -   &  3.27   & 0.36   &  2.09   & 0.27 \\
     19  &    10   &      -  &      -  &     -    &    -   &  3.14   &   0.38 &    1.99 &    0.29 \\
     20  &    10   &      -  &      -  &     -    &    -  &   2.66   & 0.33   &  2.01   & 0.29  \ML

      Mean & -      & 4.55     & 1.24    & 7.90 &   3.01 & 3.75 & 0.48 & \textbf{1.87} & \textbf{0.29} \LL \vspace{-3.0em}
 }
\end{ctable}
\endgroup

A summary of pose accuracy over all the test scenes is shown in Table~\ref{tbl:SfMcomparison}, shown in terms of the mean instantaneous pose error, and comparing the same four variants of SIFT and \gls{RLFF}-based approaches as before. We separate scenes for which all methods converged, allowing direct comparison of statistics in the top half of the table, but noting that these correspond to the easiest scenes. Overall, the monocular variant of \gls{RLFF} showed the strongest performance for these easier scenes; whereas, the stereo version performed better for the more challenging scenes, in which at least one method failed to converge.

Overall, these results show the proposed methods outperforming SIFT-based methods, and importantly, the proposed method converged in all scenes but one (mono) or two (stereo), while the SIFT-based methods failed to converge for five (mono) or eight (stereo). We see the proposed method allows SfM to operate in scenes where it previously could not, but also that it improves performance in camera pose estimation wherever these is refracted scene content, even for less challenging scenes. 
%!TEX root = main.tex

\section{Conclusions}
\label{sec:conclusions}

In conclusion, we proposed a novel 4D feature defined by the rays of light travelling through curved \glspl{RO}, as opposed to the conventional 2D image features defined by a single 3D point in space. Advantageously, our feature captures both Lambertian points and features imaged through smooth \glspl{RO}. We demonstrated methods for detecting and extracting the proposed \gls{RLFF} from LF imagery captured by a hand-held LF camera, and for employing the resulting features in conventional vision algorithms including SfM. Finally, we evaluated \gls{RLFF}'s benefits in the context of \gls{SfM}, comparing to conventional SIFT-based methods. We showed the proposed method allowed \gls{SfM} to operate where it could previously could not. We also show improved 3D reconstruction performance and 3D camera trajectory estimation. Our method is especially advantageous in scenes dominated by \gls{RO}s, in which traditional approaches fail.

For future work, we intend to develop a feature detector and descriptor that directly employ the local 4D structure of refracted features, as in the LiFF feature for Lambertian LFs~\cite{dansereau2019liff}. Demonstrating \gls{RLFF} in more scenarios is also of interest, including closed-loop control for visual servoing, and place recognition for localisation and SLAM. Finally, it is well understood that reflection off smooth curved surfaces exhibits characteristics similar to refraction through transparent objects. We expect generalisation of the \gls{RLFF} to reflective scenes to be straightforward and to show similar performance advantages as in the refractive case.

\section*{Acknowledgment}

The authors thank Steve Martin and Gavin Suddrey for helping build the camera mount and maintain the robot. We also thank the other members of the Australian Centre for Robotic Vision for their insight, support and guidance.

\ifCLASSOPTIONcaptionsoff
  \newpage
\fi



\bibliographystyle{IEEEtran}
\bibliography{IEEEabrv,bib/tsai2020rlff.bib}
\end{document}